\documentclass[sigconf,nonacm]{acmart}
\AtBeginDocument{%
  }

\usepackage[ruled,vlined]{algorithm2e} 

\begin{document}

\title[End-to-End Autonomous Generation of Assembly Plans]{End-to-End Autonomous Generation of Human Assembly Plans}

\author{Faustin Arion von Arx}
\email{fvonarx@ethz.ch}
\affiliation{%
  \institution{ETH Zurich}
  \department{Department of Mechanical and Process Engineering}
  \city{Zurich}
  \country{Switzerland}
}

\author{Millicent Schlafly}
\correspondingauthor
\email{mschlafly@ethz.ch}
\affiliation{%
  \institution{ETH Zurich}
  \department{Department of Mechanical and Process Engineering}
  \city{Zurich}
  \country{Switzerland}
}

\author{Mark Fuge}
\email{mafuge@ethz.ch} 
\affiliation{%
  \institution{ETH Zurich}
  \department{Department of Mechanical and Process Engineering}
  \city{Zurich}
  \country{Switzerland}
}


\begin{abstract}

Turning a CAD design into an assembly plan is still largely done by hand, requiring engineers to reason about geometric feasibility, tool access, stability, and the ergonomics of human assembly. In this work, we encode long-established design for assembly (DfA) principles into a contained, end-to-end approach for generating assembly plans. Our approach takes only a mesh assembly and produces either a step-by-step assembly manual or a structured failure report, requiring no joint metadata, fastener annotations, or additional information. Four major components of a manufacturing plan are addressed autonomously: an assembly tool list, the assembly sequence and subassemblies, an assembly manual, and design feedback for improving assemblability. For determining the sequence plan, we systematically disassemble the object in a physics simulator and apply a cost function that encodes DfA principles. Manual generation, tool labelling, and assembly feedback rely primarily on multimodal large language models. Compared with a baseline that always removes the outermost part first from \citet{tian2024asap}, DfA-aware sequence planning reduces simulated assembly time, measured with a robot-arm assembly-time proxy, by 35\% on 136 assemblies of 5 to 30 parts. The correct tool is selected for 88.6\% of assembly steps. A vision-language model judge compares the generated manuals against ablated variants, identifying which page elements carry the information a reader needs. The presented approach and open-source code are available for use by engineers or AI agents looking to rapidly accelerate the creation of manufacturing plans for a given product design. 
  
\end{abstract}




\maketitle

\section{Introduction}
\label{sec:introduction}

Through Design for Assembly (DfA), Boothroyd shows that assembly time can be reduced by up to 85\%~\cite{boothroyd2010product}, resulting in cost savings. These changes are currently implemented manually by human engineers over months. We seek to reduce product development and assembly time by producing a fully autonomous tool for generating human assembly plans. One key insight of this work is that we can obtain quantitative representations of assemblability for iterative optimisation by making a few core assumptions about Newtonian physics (\textit{e.g.}, gravity and geometric collisions) and incorporating widely documented DfA principles through a combination of changes to the objective terms and multimodal large language models (MLLMs). Enabled by quantitative assessments of assemblability, we address the following research questions: \textit{How can we use recent advances in MLLMs and robotics to produce a fully autonomous tool for generating human assembly plans using only component geometries?}; and \textit{To what extent does the automated optimisation using design for assembly principles reduce assembly time?} Our open-source tool could be used in a closed loop for a human engineer or AI agent to make updates to the product design.

\begin{figure*}[ht]
  \centering
  \includegraphics[width=\linewidth]{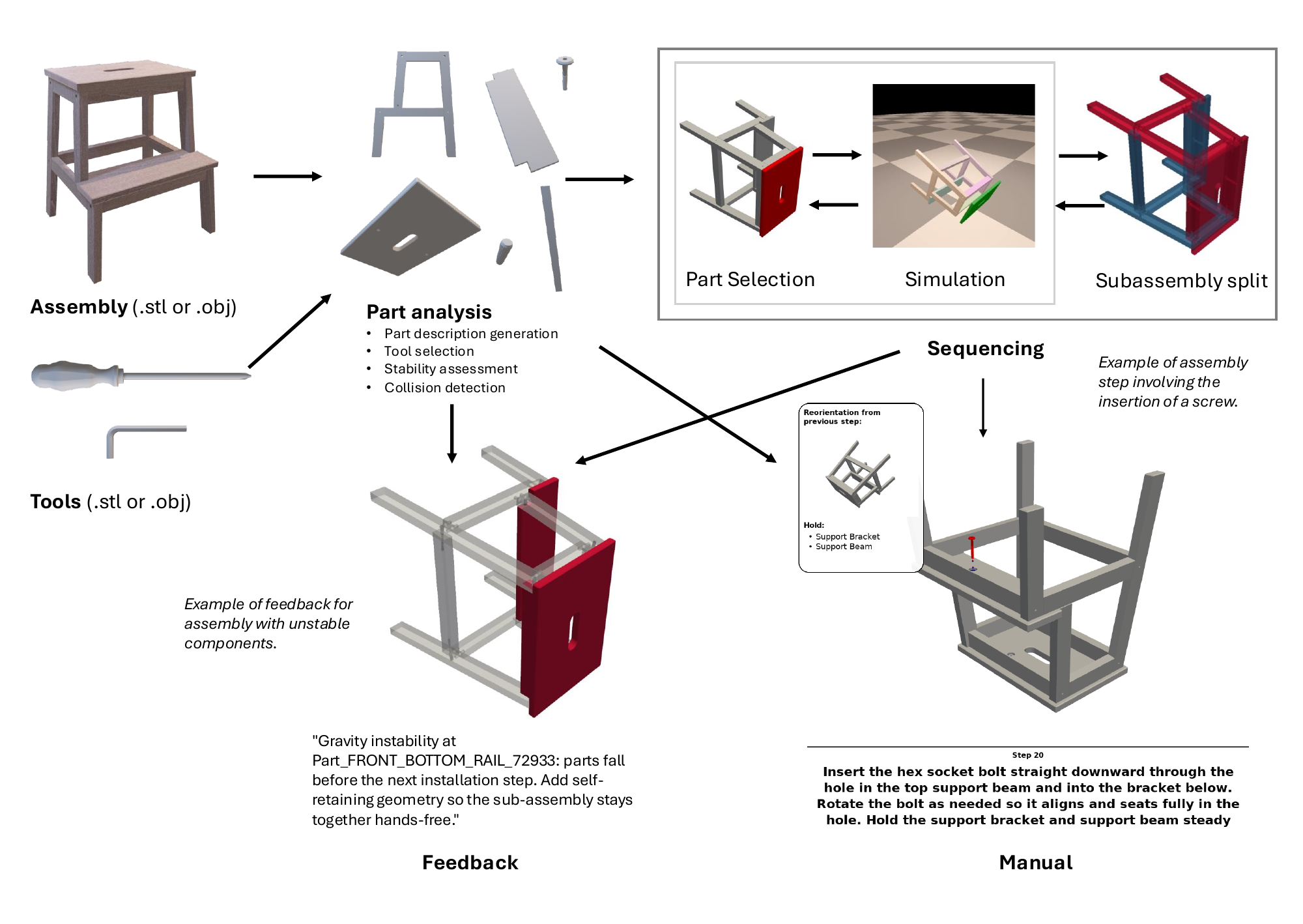}
  \caption{Overview of the pipeline, shown on the IKEA step stool from the IKEA
  3D Assembly Dataset~\cite{ikea3dassembly}. The only inputs are the part meshes
  of the assembly and of the available tools (\texttt{.stl} or \texttt{.obj}).
  \emph{Part analysis} checks the input and gathers per-part information: it
  detects overlaps between parts and tests whether the assembly is stable under
  gravity (Section~\ref{sec:methods:preprocessing}), names each part
  (Section~\ref{sec:methods:naming}), and selects the tool each part requires
  (Section~\ref{sec:methods:tools}). \emph{Sequencing} finds an assembly order by
  disassembly (Section~\ref{sec:methods:sequencing}): candidate parts are ranked
  by the DfA cost function, whose weights are tuned offline against an
  assembly-time proxy
  (Sections~\ref{sec:methods:search}--\ref{sec:methods:weight_optim}); each
  removal is checked in the physics simulator for a collision-free path, tool
  access and stability (Section~\ref{sec:methods:feasibility}); and the blocking
  information recorded during these checks is reused to split the sequence into
  subassemblies (Section~\ref{sec:methods:subassembly}). If a sequence is found,
  each step becomes a manual page with a render, a text instruction, and callouts
  for reorientation and parts to hold (Section~\ref{sec:methods:manual}); shown
  here is the insertion of a bolt. Defects found in part analysis, or a search
  that finds no sequence, are instead returned as design feedback that names the
  affected parts and suggests a change; for a buildable assembly, the same stage
  suggests changes that would make it easier to assemble
  (Section~\ref{sec:methods:feedback}). Shown here is an instability, where parts
  fall before the next step.}
  \Description{Flow diagram of the assembly-planning pipeline applied to an IKEA
  step stool. Assembly and tool meshes feed a part-analysis stage (part
  description, tool selection, stability assessment, collision detection), then a
  sequencing stage looping over part selection, simulation, and subassembly
  splitting, and finally two outputs: a design-feedback report and a step-by-step
  assembly manual.}
    \label{fig:example}
\end{figure*}

To generate our end-to-end autonomous tool, we build on prior work in generative AI and sequencing for robot assembly. 
Generative Text- and Image-to-CAD models can produce mechanical components and
articulated assemblies directly from natural language or image prompts~\cite{li2025cadllamaleveraginglargelanguage,
shui2026articadarticulatedcadassembly}. 
However, the spatial-reasoning limits of modern multimodal large
language models (MLLMs)~\cite{picard2025concept} can lead to outputs whose parts
interpenetrate, have no stable resting pose, or cannot be assembled in any
order~\cite{dong2026muse}. Complementing the capabilities of MLLMs, methods for automatic assembly sequencing offer explicit management of geometric
and combinatorial problems inherent to increasingly complicated assemblies. These approaches range from physics-based assembly-by-disassembly, which search for collision-free removal orders that remain stable under gravity~\cite{tian2022assemble, tian2024asap}, to graph- and matrix-based formulations that partition a product into feasible subassemblies and optimise the resulting sequence~\cite{demello1990, wilson1994, hartmann2025unifiedframeworkautomatedassembly}, and more recently to learned sequencing policies~\cite{shu2024s2a, wang2025alternativeplans}. However, existing pipelines expect rich
structured input---\textit{e.g.}, joint metadata, material properties, fastener annotations, or
a manually constructed liaison graph---that requires additional human annotation~\cite{jimenez2013survey}. The same gap applies to the downstream artefacts: automatic generation of assembly instructions has been studied since \citet{agrawala2003}, but existing systems likewise start from an annotated model~\cite{parzeller2026aissembly}. By combining recent advances in both fields, we provide the first end-to-end approach for generating assembly plans---including an assembly tool list, assembly sequencing, subassembly identification, design feedback, and a usable assembly manual shown for an exemplary assembly in Figure~\ref{fig:example}---with only mesh information.

Mesh-only input is aimed primarily at assemblies produced by generative models,
which output geometry without mates, joints, or fastener annotations and come
without a manual. The generated manual allows humans to stay in the loop where
needed, especially during prototyping, when a design may have to be built and
verified outside the automated pipeline. For
human-designed assemblies aimed at industrial production, mates and fasteners are
usually defined in the CAD assembly, and planners that take annotated input
directly. Human-designed models without these annotations exist nonetheless. In
master modelling, for instance, a product is designed as a single multi-body part
whose bodies are only later split into separate parts, so no mates exist until
someone builds a CAD assembly from them by hand; the same holds for models
exchanged as STL or OBJ files. We take meshes as input because nearly every 3D
representation can be converted to a mesh. Although meshes from conversion or generation are
often imperfect, we repair small surface defects with existing libraries, and
the simulator tolerates small interpenetrations, such as a slightly oversized
screw (Section~\ref{sec:methods:preprocessing}).

The contributions of this paper are as follows:
\begin{itemize}
    \item a method for identifying the necessary assembly tools for parts with a radially symmetric contact region (e.g., screwdriver and hex key) with 88.6\% accuracy over 158 parts, each added in its own assembly step;
    \item the incorporation of DfA principles into assembly sequence planning through the objective function, which reduces the estimated assembly time by 35\% on 136 assemblies of 5 to 30 parts compared to an sequentially selecting the outmost part for assembly~\cite{tian2024asap};
    \item the measurement of the effect of subassembly detection via a tree search and a DfA objective function on assembly time, demonstrating that splitting into subassemblies does not by itself reduce the estimated assembly time, and costs time with a single worker, but allows two workers to build the subassemblies in parallel, with an estimated reduction of 37\% over the same baseline, on par with the linear sequence;
    \item the use of MLLMs to create a structured manual or failure report of the assembly plan, where generated manuals are compared to ablated variants by a VLM judge; and
    \item open-source code that can be used on any set of object meshes.\footnote{https://github.com/IDEALLab/AssembleX} 
\end{itemize}

The remainder of this paper
is organised as follows. Section~\ref{sec:related_work} reviews relevant work
in automatic assembly sequencing (AAS), DfA, tool and accessibility reasoning,
assembly instruction generation, and MLLM-based geometric reasoning. Section~\ref{sec:methods}
describes the pipeline in detail, covering preprocessing and geometry analysis,
sequence and subassembly finding, and manual and feedback generation.
Section~\ref{sec:results} evaluates the pipeline component-wise: tool
application on the IKEA 3D Assembly Dataset~\cite{ikea3dassembly}, and sequence
planning on the assembly dataset of \citet{tian2024asap}. Section~\ref{sec:conclusion}
discusses the findings, summarises the contributions, and outlines directions
for future research.

\section{Related Work}
\label{sec:related_work}

The pipeline proposed in this work sits at the intersection of several
established research areas: AAS, DfA, the geometric reasoning needed to place
tools and fasteners, the presentation of assembly instructions to a human, and
the application of MLLMs to CAD understanding. Each has a substantial literature
of its own; this section reviews the parts most directly relevant to the
subproblems addressed in Section~\ref{sec:methods}, and, for each, the input
requirements that separate the existing methods from the mesh-only setting
targeted here.

\subsection{Automatic Assembly Sequencing}
\label{sec:related:asp}

Automatic assembly sequencing has historically been formulated along two
complementary axes: as a feasibility problem, concerned with finding any valid
order in which a given set of parts can be put together; and as an optimisation
problem, concerned with selecting from among the feasible sequences one that
minimises a cost such as assembly time, robot motion, or fixturing complexity.
The classical feasibility formulations reduce the $O(n!)$ space of orderings to
a set of answerable questions about part liaisons~\cite{defazio1987} and encode
the admissible orders compactly as an AND/OR graph over
subassemblies~\cite{demello1990andor,demello1990}. \citet{jimenez2013survey}
surveys the combinatorial and geometric variants of the problem, and \citet{ghandi2015review} give a taxonomy of the assembly and disassembly
path-planning methods used to answer the underlying geometric queries, including
the metaheuristic solvers that dominate the optimisation-oriented literature. The
same search structure underlies disassembly sequencing for maintenance and
end-of-life recovery~\cite{lambert2003disassembly}, which optimises a different
objective over the same feasibility constraints.

What most of these methods share is a dependence on a liaison graph, precedence
relations, or interference matrices supplied per assembly. Physics-based
planning removes that dependence by evaluating feasibility in simulation
instead. \citet{tian2022assemble} search for collision-free
disassembly paths for general part geometry, and their subsequent
work~\cite{tian2024asap}, on which the sequence-finding stage of our pipeline
builds most directly, extends this to full sequences: for a given assembly
state, a physics simulator determines which parts can be removed without
collision and whether the remainder stays stable under gravity. The approach
orders candidate parts only to reach a feasible sequence quickly and does not
optimise any measure of sequence quality. Fabrica~\cite{tian2025fabrica} carries the same research direction
through to physical execution by adding grasp, motion, and fixture planning for
a dual-arm robot. Similarly, \citet{wan2018assembly} combine stability, quasistatic graspability, and geometric assemblability constraints to produce sequences directly suited for motion planners. \citet{huang2021newanalog} show, for robotic
construction, how design intent can be linked to the assembly planner so that
sequencing constraints are visible while the structure is still being designed.
For scaffold- and fixture-free assembly of slender frame and truss structures,
the elastic deformation of the partial structure can itself serve as the
sequencing objective: \citet{wang2023frameassembly} optimise the
sequence against this deformation as a temporally coherent topology optimisation,
and \citet{huang2024constructability} compare design alternatives by
the score of their optimised sequences, a form of DfA practice close to the
feedback stage of Section~\ref{sec:methods:feedback}.
\citet{hartmann2025unifiedframeworkautomatedassembly} couple
feasibility checks with cost-based sequence selection, but, like most
optimisation-oriented work in this area, assume that the input assembly is
annotated with joint, fastener, or contact information that is not present in
raw mesh output. The same reliance appears where physical validity is used as a
design constraint rather than a sequencing one: FabHacks~\cite{mei2024fabhacks}
searches over ways of connecting existing parts and settles the result with a
physics solver, but draws its candidate connections from a catalogue of
hand-specified connector types.

A recent line replaces the search with a learned policy. \citet{shu2024s2a} train a graph attention network over the assembly graph
with delayed rewards, \citet{vallat2023rl} learn a sequencing policy
under a stability constraint for scaffold-free construction, and \citet{wang2025alternativeplans} combine assembly-by-disassembly with a graph
network and a GPU stability simulator to generate alternative plans online. These
methods amortise the cost of search across many assemblies, at the price of a
training set and of a policy whose objective is fixed at training time. We keep
an explicit search for two reasons: no dataset pairs assembly sequences with
measured human assembly times, so there is no supervision for the quantity we
optimise; and an interpretable cost function lets each term be traced back to a
DfA principle and reweighted (Section~\ref{sec:methods:cost}).

\subsection{Subassembly Identification}
Subassembly identification has been studied extensively as a means of reducing
the complexity of assembly sequence planning and enabling parallel assembly
workstreams~\cite{bahuba2019}. Early work by \citet{demello1990} uses a \textit{cut-set method} on the assembly's
liaison graph: edges are cut to partition the graph into two connected
components, each corresponding to a candidate subassembly. The method is
complete but requires a separate geometric feasibility test for every
candidate, which becomes intractable for large assemblies.

\citet{wilson1994} address this intractability by introducing the
\textit{Non-Directional Blocking Graph} (NDBG), a polynomial-space
representation of pairwise part blocking computed directly from CAD geometry,
which enables efficient identification of directionally removable
subassemblies; they also prove that the connected variant of the partitioning
problem is NP-complete, motivating heuristic search. \citet{watson2019} take a related geometric approach with their
\textit{Disassembly Interference Graph} (DIG) but treat blocking as a soft
cost to minimise rather than a hard constraint, producing a hierarchical
decomposition tree that exposes parallelism. A separate line of work focuses on
subassembly \textit{stability}: \citet{bahuba2019} and
\citet{gulivindala2020} classify contacts by stability type
and use these predicates to filter candidates without simulation, at the cost
of limited applicability to non-standard contact geometries. The foundational
matrix-based approach of \citet{Dini1992} predates these
methods but relies on liaison and precedence matrices that must be provided
separately for each assembly, again precluding application to raw geometric
input.

What unites these methods is a reliance on geometric computation or matrix
construction performed as a dedicated, standalone stage. The subassembly
detection approach proposed in Section~\ref{sec:methods:subassembly} differs in
that it derives an approximate directional blocking matrix as a by-product of
the sequence-finding stage, recovering most of the structural benefit of NDBG-
and DIG-style methods at negligible additional cost.

\subsection{Design for Assembly and Assembly Time Estimation}
\label{sec:related:dfa}
Design for Assembly is concerned with shaping product designs so that they are
easy and fast to assemble by either human or robot. Classical DfA, formalised
by \citet{boothroyd1994,boothroyd2010product}, contributes quantitative
scores for design decisions that have been empirically validated as predictors of
human assembly time. Common principles include minimising part count, preferring
symmetric parts, working top-down, and avoiding reorientations. The same
principles remain the basis of industrial design for manufacturing practice, where the principles are applied
by hand to an existing design to recover part count and cost
reductions~\cite{butt2020} of the kind this work seeks to produce automatically.
More recently, \citet{picard2025concept} evaluate vision-language
models (VLMs) on assessing designs against such criteria directly from visual
input, sidestepping the need for explicit feature annotation.

DfA scores are one of two traditions for quantifying manual assembly
effort. The other comprises the predetermined motion time systems used in industrial
engineering, which decompose a task into elementary motions --- reach, grasp,
move, position --- and sum tabulated times for each~\cite{maynard1948mtm}.
These systems supply the ground truth that DfA scores approximate, but they
take a described motion sequence rather than geometry as input. Inside a sequence
search, the insertion path of each step is already computed during state
validation, so a model that maps such a path to the elementary motions of a
human assembler could in principle make these systems usable there; no such
mapping exists yet, and learning one would require motion-level human data that
is not available for the assemblies considered here. Their accuracy against
measured times is also imperfect, and the deviations are structured by movement distance and
placement precision in the manner predicted by Fitts's
law~\cite{fitts1954,mazareinezhad2025most}. Both observations motivate the
design of our cost function (Section~\ref{sec:methods:cost}): each term
corresponds to a DfA principle for which there is independent evidence of an
effect on human assembly time and which can be evaluated from quantities the
search already computes, and the use of such heuristics as a proxy for assembly
time in the absence of labelled data is supported by prior
work~\cite{jiang1994optimization}.

\subsection{Tool, Fastener, and Accessibility Reasoning}
\label{sec:related:tools}
Deciding that a part needs a tool, and that the tool can physically reach it, is
a geometric problem that the assembly-planning literature has treated separately
from sequencing. \citet{wilson1996tools} formalises it through a
\textit{use volume}, the space that must remain free for a tool to be applied,
together with placement constraints relating that volume to the parts, which
turns tool applicability into an interference query of the same kind used for
part removal. \citet{kang2008tool} extend the check to cover both
tool motion and tool placement during assembly planning. Both approaches
presuppose that the fastener and its associated tool are already identified in
the model.

Supplying that missing annotation from geometry alone is the subject of a
separate line of work. Feature recognition maps CAD geometry to manufacturing
features, either through rules over the boundary representation or, more
recently, through networks trained on labelled solids~\cite{yeo2021feature}, and
mate and joint prediction learns which faces of two parts are meant to be
joined and how~\cite{jones2021automate,willis2021}. Related work in
computational fabrication treats the joint itself as the design variable rather
than as an input, optimising its shape for stiffness under a differentiable
contact model~\cite{sun2023dovetail}. These methods depend on parametric B-rep
input with joint or mate supervision, which a raw mesh does not carry. The tool
stage described in Section~\ref{sec:methods:tools} instead recovers candidate
application axes from a convex decomposition of the tool mesh and defers the
semantic decision to a VLM, then validates the result with an accessibility
check in the spirit of \citet{wilson1996tools}.

Accessibility has also been computed directly from geometry in inspection,
machining, and moulding. For, example, the directions from which a probe or cutter can reach a
surface point form an accessibility cone~\cite{spyridi1990accessibility,
spitz1999accessibility, dhaliwal2003cones}. Similar approaches have been developed in moulding~\cite{chen1993parting, ahn2002cast} and milling \citet{elber1994milling}. Graphics hardware makes accessibility computations fast enough
to sit inside a planner~\cite{spitz2000graphics, khardekar2006mold}, and learned
predictors now approximate them for arbitrary cutter
geometry~\cite{zhong2025deepmill}. These approaches could complement our work here, helping discard candidate axes before the options are presented to the VLM and address a failure mode observed on tools
with many candidate axes
(Appendix~\ref{sec:appendix:contact})


\subsection{Generating Assembly Instructions}
\label{sec:related:instructions}
Producing the sequence is only half of a usable assembly plan; the other half is
presenting it. \citet{agrawala2003} establishes the reference
treatment of the presentation problem, deriving a set of design principles for
step-by-step assembly diagrams --- step granularity, part highlighting, action
arrows, and viewpoint selection --- from studies of how people draw and read
such instructions, and building a system that lays out diagrams accordingly.
\citet{heiser2004} validate those principles against human
performance. \citet{michas2000} show that combining image and
text conveys a procedure more effectively than either alone, and later work
finds that the delivery medium changes assembly time and error rate
independently of content, with augmented-reality guidance outperforming 2D
documents across skill levels~\cite{marino2024ar} and XR registration to the
physical workpiece extending this to large structures~\cite{moisi2025digitaltwins}.
These results inform the fixed page layout adopted in
Section~\ref{sec:methods:manual}.

A recent body of work runs the pipeline in the opposite direction, taking an
existing manual as input: IKEA-Manual~\cite{wang2022ikeamanual} pairs annotated
manuals with 3D parts, its extension grounds the steps in assembly
video~\cite{liu2024ikeaatwork}, and other work parses a visual manual into an
executable plan~\cite{wang2022lego} or into part poses and
orderings~\cite{zhang2025manualpa}, up to driving a robot directly from the
manual with a VLM~\cite{tie2025manual2skill}. Generating rather than consuming
instructions has received less attention. \citet{widulle2024instructions} pass a reinforcement-learned assembly
sequence to an LLM that writes the instruction text, and
AIssembly~\cite{parzeller2026aissembly} generates work instructions from CAD by
disassembly, closest in spirit to our manual stage but requiring the CAD model
to be enriched beforehand with assembly-relevant, non-geometric information.

\subsection{Multimodal Models for CAD, Design, and Evaluation}
\label{sec:related:mllm}
Our pipeline delegates the perception and judgement steps that are not purely
geometric to a VLM (naming each part, deciding which tool it requires, and
assessing a design for assemblability), so its reliability depends on how well
current models reason about CAD geometry and spatial structure from visual
input. \citet{Khan2025} quantify VLM accuracy at recognising
manufacturing features in CAD models and identify prompt-engineering
techniques (multi-view rendering, chain-of-thought, and zero- versus few-shot
prompting) that change recognition accuracy. \citet{picard2025concept} evaluate VLMs on judging engineering designs
against manufacturability and assembly criteria directly from renders, and the
DesignQA benchmark of \citet{doris2024designqa} finds that current
MLLMs struggle to recognise technical components in CAD images and to reason
over engineering drawings. Benchmarks aimed at spatial reasoning locate the
weakness more precisely: performance degrades on rotation, perspective-taking,
and occlusion~\cite{jia2026omnispatial}, which are the operations our
axis-selection and tool-decision prompts depend on, and which motivate the
multi-view renders used throughout Section~\ref{sec:methods}.

The same models are now used to generate CAD rather than only to read it.
Text-to-CAD methods produce parametric construction sequences from natural
language~\cite{khan2024text2cad,li2025cadllamaleveraginglargelanguage} and
articulated assemblies through multi-agent code
generation~\cite{shui2026articadarticulatedcadassembly}, but their outputs are
not guaranteed to be manufacturable or assemblable~\cite{dong2026muse}. This has
prompted architectures that place an explicit verifier in the generation
loop~\cite{berger2026physicsinloop}; the pipeline presented here is intended to
serve as such a verifier for the assembly stage specifically.

Both the manual evaluation and the tool decision are scored by a model rather
than a human. LLM- and VLM-as-judge protocols have been characterised in
detail~\cite{gu2025surveyllmasajudge}, and agreement with human raters is
highest when the judging task is binary and tightly
constrained~\cite{han2025judgesverdictcomprehensiveanalysis,chen2024mllmasajudgeassessingmultimodalllmasajudge,lee2024prometheusvisionvisionlanguagemodeljudge},
which is the regime the protocol of Section~\ref{sec:results:manual} operates in.

\section{Methods}
\label{sec:methods}

The pipeline takes a CAD assembly given as a set of part meshes and produces
either a step-by-step assembly manual, when the assembly is feasible, or
targeted design feedback, when it is not. It is organised into four stages,
shown schematically in Figure~\ref{fig:example}: preprocessing of the input
geometry, inference of part-level information such as
part names and required tools, search for an assembly sequence optimised for
human assembly time, and generation of the manual or feedback from the
resulting sequence. Each stage is covered in one of the subsections that
follow.

\subsection{Preprocessing}
\label{sec:methods:preprocessing}

The pipeline accepts an assembly as a set of part meshes in a common global
frame. Generative CAD models routinely produce meshes with defects that the
downstream physics simulator would fail on or silently mishandle, so three
classes are handled first and, where relevant, forwarded to the feedback stage.
Non-watertight surfaces, introduced when generative STEP output is converted to
mesh, are re-closed with existing libraries such as Trimesh. Overlaps between
parts are detected but not repaired. Small interpenetrations, such as a slightly
oversized screw, are absorbed by the simulator. Larger overlaps, which are
frequent in AI-generated meshes~\cite{gao2025}, are reported to the feedback
stage as a design defect rather than removed by moving the parts, since moving a
part could change the function of the assembly. Finally, generative outputs often lack a stable
resting configuration~\cite{dong2026muse}: the assembly is dropped under gravity,
any part that moves relative to the rest is flagged, and flagged parts are
iteratively joined to their neighbours until a minimal set of fastening relations
holds the assembly together. This stability check builds on the open-source code of \citet{tian2024asap}, with its instability criterion reworked substantially for
the inputs seen here and reused during sequence search
(Section~\ref{sec:methods:feasibility}).

\subsection{Geometry Analysis}
\label{sec:methods:geometry}

\subsubsection{Naming parts}
\label{sec:methods:naming}
Each part is shown to a VLM in isolation and in the context of the surrounding
assembly, and the VLM returns a short description of the part and its likely
function that serves as a consistent handle in every downstream prompt, avoiding
repeated re-rendering. To improve reliability, the prompt combines the
multi-view, chain-of-thought, and zero-shot techniques of \citet{Khan2025}, which are reused across almost every VLM prompt in the
pipeline, including tool selection.

\subsubsection{Tool Application}
\label{sec:methods:tools}
For each part, the pipeline decides whether a tool is required and, if so, which
one from the supplied catalogue and how it applies. Tools are supplied as
meshes, like the assembly parts, so their application axis has to be recovered
from geometry alone; the renders used for this are shown in
Figure~\ref{fig:appendix:tool_axis} in the appendix. The contact region of
a screwing tool (a screwdriver bit, wrench socket, or hex driver tip) is
approximately radially symmetric about its application axis. When the tool mesh
is split into a convex decomposition, that axis therefore appears as the minor
principal axis of one of the resulting components, and the minor principal axes
of all components form a short list of candidate application axes. The
candidates are drawn as numbered arrows on a render of the tool, and a VLM
selects the application axis (Figure~\ref{fig:appendix:tool_axis}a). In a second
pass, the VLM picks the end of that axis at which the tool meets the part, which
fixes the contact point and the application direction
(Figure~\ref{fig:appendix:tool_axis}b). Tools without a radially symmetric
contact region fall outside this assumption. To decide whether a part needs a
tool at all, a render of each candidate tool applied to the part is passed to
the VLM, which judges whether a tool is needed. The chosen tool is then placed
at the contact point during sequence finding and checked for collisions with the
rest of the assembly (Section~\ref{sec:methods:feasibility}), which is the
mesh-only analogue of the use-volume accessibility test of
\citet{wilson1996tools}.

Fasteners are ordinary part meshes, present in the input like the rest of the
assembly, as they are in the IKEA models, and they are not treated differently
during sequencing. A screw or dowel is named by the VLM like any other part
(Section~\ref{sec:methods:naming}), and the tool decision above assigns it the
tool it needs; no fastener annotation or joint type is required. Like every
other part, each fastener is added in a step of its own at the point the search
places it, which is why fasteners appear in the generated manuals despite the
restriction to single-part steps (Section~\ref{sec:limitations}); any part that
would fall at that point is counted as a part to be held
($C_{\text{hold}}$, Section~\ref{sec:methods:cost}). Fasteners missing from the
input are not inferred; a joint that cannot hold without one appears instead as
an instability, for which preprocessing suggests an added fastening relation
(Section~\ref{sec:methods:preprocessing}).

\subsection{Sequencing}
\label{sec:methods:sequencing}
The number of possible orderings for an assembly of $n$ parts scales as
$O(n!)$, so the naive search space becomes intractable even for small
assemblies. We limit it by considering parts one at a time, pruning physically
infeasible candidates (Section~\ref{sec:methods:feasibility}) and ranking the
survivors against a DfA cost function (Section~\ref{sec:methods:cost}), whose
term weights are tuned to an assembly-time proxy in the absence of
labelled human data (Section~\ref{sec:methods:weight_optim}).
The sequence produced by this linear, part-by-part search is then revisited in a
separate pass to
identify groups of parts that can be assembled in parallel
(Section~\ref{sec:methods:subassembly}). The end-to-end procedure is summarised in Algorithm~\ref{alg:appendix:sequence_planner} in the appendix.

\subsubsection{Feasibility Metric}
\label{sec:methods:feasibility}
Two terms are used throughout this section: \emph{path finding} refers to
determining the trajectory a single part takes to enter or leave the assembly,
and \emph{sequence finding} to determining the order in which the parts are
added. A third term, the \emph{contact graph}, is referenced repeatedly:
each part is a node, and an edge joins two parts whose surfaces lie within a
small tolerance of one another. To check whether a single part can be removed and, if so, to find a
valid path, we use the assembly-by-disassembly code released by \citet{tian2024asap}, which also supplies the gravity-based stability test.

A removal step is therefore physically feasible if and only if three criteria
hold:
\begin{enumerate}
  \item There exists a path along which the candidate part can be removed while
    all others are fixed.
  \item If the part requires a tool, the tool can be applied and moved into
    position without collision.
  \item The resulting subassembly is globally and locally stable once at most two
    parts are held in place; the parts that must be held are recorded and enter
    the cost through $C_{\text{hold}}$ (Section~\ref{sec:methods:cost}).
\end{enumerate}

For criteria 1 and 3, the simulator is built on the open-source codebase of \citet{tian2024asap}. Two changes are necessary to make it usable here. Both
checks are parallelised across candidates, since the original sequential
version is too slow to run at every node of the search. The stability check itself is also
reworked to assess the movement of an individual part relative to the whole assembly. Now, a part fails the stability check only
when it moves relative to the bulk, not when the assembly as a whole is
perturbed; an outlier sliding out of an otherwise rigid assembly is detected,
while a uniform drop or rotation is ignored. The check does not fix the global
orientation of the assembly. Three stable resting poses of the assembly, treated
as a single rigid body, are computed, and the assembly is tested under gravity in
each of them, so that parts falling out are detected even when the assembly as a
whole rests stably. A step passes if the assembly is stable in at least one of
these poses. This orientation handling is taken from the code of \citet{tian2024asap}, adapted to the stability criterion above. Rotating
the assembly between steps is allowed and penalised by the reorientation term
$C_{\text{pose}}$, while $C_{\text{top}}$ rewards inserting the next part from
above (Section~\ref{sec:methods:cost}), which mirrors a human assembler turning
the partial assembly so that the next part goes in top-down.

\subsubsection{Search Problem}
\label{sec:methods:search}

The search is a beam search: all removal candidates of the current frontier are
checked in one parallel batch, and the top-$k$ partial sequences by cost
(Section~\ref{sec:methods:cost}) are kept and expanded at the next step; a
larger $k$ trades more simulation per frontier for lower-cost sequences. We use
$k = 4$. A guaranteed global optimum would require exploring the full tree, whereas
a narrow beam already finds a sequence close to the optimum in a fraction of that
time (Appendix~\ref{sec:appendix:beam_search}). The number of
candidate checks per plan is capped at 6{,}000. Once the search has finished,
the lowest-cost complete sequence in the explored tree is returned, rather than
the one reached by following the best-ranked branch: since the states of the
tree are the sets of parts still assembled, this is a shortest-path query over
the removal steps already checked, and it requires no further simulation.

\subsubsection{Cost Function}
\label{sec:methods:cost}
The quantity being optimised is the time it would take a human to perform the
assembly, approximated in practice by the proxy of
Section~\ref{sec:methods:weight_optim}. No public dataset associates sequences
with assembly times, so we instead use a weighted combination of factors that are
empirically known to reduce assembly time and that can be evaluated cheaply from
the information available during search.

The cumulative cost up to step $i$ is the sum of per-step costs, each of which
is a weighted combination of five terms:

\begin{align}
  J_{i} &= \sum_{j=0}^{i} C(x_j), \\
  C(x_{i}) &= w_{\text{pose}} C_{\text{pose}}(x_i)
            + w_{\text{top}} C_{\text{top}}(x_i) \nonumber \\
           &\quad + w_{\text{dist}} C_{\text{dist}}(x_i)
            + w_{\text{free}} C_{\text{free}}(x_i) \nonumber \\
           &\quad + w_{\text{hold}} C_{\text{hold}}(x_i)
\end{align}

The individual costs are defined in Table~\ref{tab:methods:cost_functions}.

\begin{table*}[t]
  \centering
  \caption{Assembly sequencing cost functions and their weights. Each principle is found in the design for assembly literature. The reference weights $w_{\text{ref}}$ are set by hand; the optimised weights $w_{\text{opt}}$ are fitted to the assembly-time proxy (Section~\ref{sec:methods:weight_optim}). Note that the magnitudes of the five metrics are not directly interpretable as a measure of
  each term's influence on the chosen sequence or as a measure of real-world significance.}
  \label{tab:methods:cost_functions}
  \begin{tabular}{@{}p{0.17\textwidth} p{0.34\textwidth} p{0.28\textwidth} c c@{}}
    \toprule
    \textbf{Principle} & \textbf{Metric} & \textbf{Expression} & \textbf{$w_{\text{ref}}$} & \textbf{$w_{\text{opt}}$} \\
    \midrule
    \textbf{Minimise \newline reorientations} & Pose change since the last step; rotating the assembly between steps adds setup time~\cite{boothroyd2010product}. & $\begin{aligned} C_{\text{pose}} &= \tfrac{1 - \cos\theta}{2} \\ \cos\theta &= \tfrac{\operatorname{tr}(R_{\text{parent}}^\top R_{\text{cur}}) - 1}{2} \end{aligned}$ & 0.67 & 51.1 \\
    \midrule
    \textbf{Top-down} & Path alignment with the global $z$-axis; working top-down keeps gravity in the assembler's favour and improves visibility of the working area~\cite{boothroyd2010product}. & $C_{\text{top}} = 1 - \left(\hat{d}_n \cdot \hat{z}_{\text{down}}\right)$ & 0.67 & 3.63 \\
    \midrule
    \textbf{Minimise \newline distance} between steps & Shortest path to a previously removed part in the contact graph; switching between distant regions wastes time, as captured by Fitts's law~\cite{boothroyd2010product, fitts1954}. & $C_{\text{dist}}(x_i) = \ln \left( \min_{v \,\in\, \mathcal{S}} \, d_G(x_i, v) + 1 \right)$ & 0.33 & 41.8 \\
    \midrule
    \textbf{Prefer free} over \newline restricted paths & Count of blocked degrees of freedom; parts with many unblocked degrees of freedom are easier to handle and, if added last, do not destabilise assembled components~\cite{boothroyd2010product}. & $C_{\text{free}} = \|\mathbf{d}\|_1$ & 0.2 & 0.072 \\
    \midrule
    \textbf{Minimise parts held} & Number of parts that would fall if not manually stabilised; supporting a part while operating on it forces the assembler to multitask and slows the step~\cite{boothroyd2010product}. & $C_{\text{hold}} = |\,\text{parts\_fix}\,|$ & 1 & 1 \\
    \bottomrule
  \end{tabular}
\end{table*}

\subsubsection{Weight Optimisation}
\label{sec:methods:weight_optim}
Tuning the weights requires a proxy for assembly time, since no human data is
available that could assign an accurate cost to a given sequence. The proxy
estimates the time a robot arm on a base that moves around the assembly needs to
carry out the sequence. It has three components: the motion of the part itself,
taken from the length of its extraction path; the travel of the base around the
assembly between consecutive steps; and the reorientation of the assembly
between steps, at 0.15\,rad/s. Parts that must be held in place add no time;
they are counted separately and reported alongside the proxy time. All proxy
times reported in Section~\ref{sec:results:sequence} use this approximation, for
every method compared there.

Before any training, we set the reference weights $w_{\text{ref}}$ in
Table~\ref{tab:methods:cost_functions} by hand, judging the manuals generated
from the resulting sequences and how intuitive their steps are to follow. These
weights are not fitted to the proxy.

We optimise four of the five cost-function weights, keeping
$w_{\text{hold}} = 1$ fixed to set the scale, with a tree-structured Parzen
estimator (TPE)~\cite{bergstra2011tpe} via the Optuna
package~\cite{akiba2019optuna}. The objective is the proxy assembly time of the
sequence the planner produces under a given weight vector, aggregated over the
training assemblies; it is discontinuous because a small change in the weights
can flip which sequence the planner selects, producing a jump in the recorded
time.

TPE is a sequential model-based optimiser for expensive black-box objectives. It
uses no gradients and, unlike a genetic algorithm, keeps no population to mutate;
instead, each proposal is informed by all trials evaluated so far. After a few
random initial trials, TPE splits the evaluated weight vectors into a good group,
the best-scoring fraction of trials, and the rest, and fits a kernel density
estimate to each group. The next weight vector is the candidate with the highest
ratio of the good-group density to the other, which is equivalent to maximising
the expected improvement over the current best~\cite{bergstra2011tpe}. Because it
needs only objective values, TPE handles the discontinuous objective above and
reaches good weights in few evaluations, each of which requires sequencing every
training assembly.

The optimisation uses the training set of six assemblies with 8 to 13 parts
introduced in Section~\ref{sec:methods:experimental_evaluation}.
Two considerations motivate the size of this set. First, the cost function has only four free weights, so the surrogate has
few parameters to identify and convergence is fast regardless of dataset size.
Second, the proxy used to evaluate the optimised planner is the same one used for
training, so a larger set assembled from weaker samples would risk fitting the
cost function to artefacts of the proxy that would then resurface as apparent
gains at test time. A small set of assemblies chosen for quality is therefore
preferable to a larger one.

This proxy captures only part of what makes a step difficult for a human; a
labelled dataset of sequences with measured human assembly times, or an
elementary-motion decomposition of the kind used by predetermined motion time
systems~\cite{maynard1948mtm, mazareinezhad2025most}, estimated from the
insertion paths the search already computes (Section~\ref{sec:related:dfa}),
would let the cost function
be supervised directly rather than against a proxy, surfacing the factors it is
blind to (precision under tight fits, part weight, handlability), which we leave
to future work.

\subsubsection{Subassembly Detection}
\label{sec:methods:subassembly}
Splitting the assembly into subassemblies that can be put together
independently before being joined is a well-established way of reducing total
assembly time~\cite{bahuba2019}. The two main obstacles to automating this step
are the combinatorial size of the candidate space and the cost of determining
which subsets are geometrically separable from the rest. We sidestep the second
cost by reusing blocking information that is already produced as a by-product of
the disassembly-path search in Section~\ref{sec:methods:feasibility}.

For the purposes of this work, a subassembly is a connected subset $S$ of the
contact graph $A$ that can be rigidly translated in at least one of the six
axis-aligned directions $\{\pm x, \pm y, \pm z\}$ without colliding with the
complementary set $A \setminus S$. Restricting to axis-aligned directions is a
deliberate simplification that makes the search tractable; subassemblies that
require an oblique removal direction or sequential translations in different directions are not identified, although future work could address this.

The procedure has three steps: building an approximate directional blocking
matrix, enumerating candidate subassemblies by tree search, and
validating the highest-scoring candidates in the physics simulator.

\paragraph{1. Approximate directional blocking matrix.}
We maintain a matrix $B[i, \mathbf{d}]$ that stores, for each part $i$ and each
of the six directions $\mathbf{d}$, the set of parts known to block $i$ in
$\mathbf{d}$, together with a flag marking $i$ as known-free in $\mathbf{d}$. This information is naturally collected during the sequencing simulation step, where part $i$ is subjected to a force in each direction $\mathbf{d}$ separately. The part is defined as free if it can be fully removed from the assembly and blocked otherwise. Note that the part is not limited to moving in the direction $\mathbf{d}$, meaning that, for example, slightly misaligned or curved parts can still be correctly classified.
Three rules populate $B$ incrementally as the disassembly proceeds:

\begin{itemize}
  \item \textbf{Freedom.} If part $i$ is observed to be free in direction
    $\mathbf{d}$, no remaining part can be blocking it: set
    $B[i, \mathbf{d}] = \emptyset$ and flag $i$ as free in $\mathbf{d}$.
  \item \textbf{Attribution.} If removing part $k$ unlocks direction
    $\mathbf{d}$ for part $i$, add $k$ to $B[i, \mathbf{d}]$.
  \item \textbf{Symmetry.} If part $j$ blocks part $i$ in direction
    $\mathbf{d}$, that is $j \in B[i, \mathbf{d}]$, then $i \in B[j, -\mathbf{d}]$,
    since contact blocking is reciprocal.
\end{itemize}

The matrix is a strict underapproximation of the true blocking relation: a
direction in which $i$ is simultaneously blocked by several parts only
attributes the last of them to be removed, since removing the first does not free $i$. The first part is thus assumed to be non-blocking --- the conservative policy of marking every unknown part as blocking would eliminate feasible subassemblies. This underapproximation, however,
never excludes a valid subassembly, since every entry in $B$ is a true blocker
and no spurious blocking edge can hide a removable subset; it only lets some
infeasible candidates slip through to the simulator in Step~3, where they are
filtered out. An example of the generated blocking matrices and the corresponding contact graph is shown in Figure~\ref{fig:appendix:blocking_matrix} in the appendix.

\paragraph{2. Tree search over connected subsets.}
Candidate subassemblies are enumerated by a bottom-up tree search. Each node
represents a connected subset $S$, extended by adding a contact-graph neighbour
of some part already in $S$, which maintains connectivity by construction. For
each candidate we track, per direction, an external blocking count
$c_{\mathbf{d}}(S)$ giving the number of parts outside $S$ known to block some
part inside $S$ in direction $\mathbf{d}$. The update upon adding a part is
incremental and only inspects the blocking entries of that part.

$S$ is recorded as a candidate whenever $c_{\mathbf{d}}(S) = 0$ for at least one
direction. A branch is pruned when no such direction remains. Because the
per-node work is negligible, exhaustive search is tractable for all but very
large assemblies. To avoid repeating the search at every disassembly step, the
set of feasible subassemblies from the previous step is reused: removing the
just-disassembled part from each candidate yields the feasible set at the
current step, including subassemblies that only become viable later in the
sequence.

\paragraph{3. Scoring and simulator validation.}
A single-part subassembly is trivially valid but provides no benefit, and a
subassembly that fragments the remainder may add fixturing complexity rather
than reduce it. We therefore score each candidate using three factors:

\begin{itemize}
  \item \textbf{Size} $|S|$, rewarding intermediate sizes;
  \item \textbf{Freeness}, the number of directions in which $S$ is externally
    unblocked, which doubles as a soft indicator that the approximation has not
    hidden a blocker;
  \item \textbf{Remainder fragmentation}, the number of connected components of
    $A \setminus S$, which is penalised.
\end{itemize}

The three factors are combined into a single score used to rank the candidates.
This score is only a proxy for the true cost function
(Section~\ref{sec:methods:cost}), which cannot be recovered from the blocking
matrix $B$ and would require rerunning the physics simulation for every
candidate. This allows for the quick elimination of a large set of feasible but unproductive subassembly candidates, such that following computationally heavier operations can be performed.

The top-$k_S$ ranked candidates are validated in the physics simulator from
Section~\ref{sec:methods:feasibility}: $S$ is rigidly translated in its
candidate removal direction and the simulator checks for collisions and for
instability of $A \setminus S$ under gravity. For each candidate that passes
validation, the parts of $S$ and of $A \setminus S$ keep the relative order
they have in the linear sequence; the two subsequences are not re-optimised. The
split's cost combines the two sub-sequence costs with that of the joining
step, and the split is accepted only if this total is lower than that of the
corresponding linear sequence over the same parts. The procedure recurses on $S$
and $A \setminus S$, producing a hierarchical decomposition tree that terminates
when no further split is beneficial or at a depth of two.

\subsection{Assembly Manual Generation}
\label{sec:methods:manual}
Given a feasible sequence and the per-step metadata produced by the search
(Section~\ref{sec:methods:sequencing}), the assembly manual stage converts each step into
a self-contained instruction page for a human assembler by laying out the
available information in a fixed visual template.

Combining image and text has been shown to yield clearer assembly instructions
than either alone and to outperform even video instructions~\cite{michas2000}:
images convey the required motion at a glance, while text disambiguates and
catches errors that an image cannot. The manual adopts this combined format.
Each page uses a fixed layout consisting of a coloured render of the action
(with the part being added highlighted against the rest of the assembly, and the
required motion marked by dots), a text instruction generated by a VLM, and two
conditional callouts: one for a reorientation relative to the previous step and
one for the required tool. Highlighting the part being added and marking its
motion follow the diagram principles of \citet{agrawala2003},
validated against human performance by \citet{heiser2004}. To maximise visibility, the assembly motion is
rendered from four evenly distributed angles, and each angle is scored by the
total pixel-wise difference between the assembled and disassembled frames. The angle with the
largest dissimilarity is chosen, since it is the view in which the part's motion
is most visible on screen. A representative output is shown in
Figure~\ref{fig:appendix:manual_example} in the appendix, with validation
discussed in Section~\ref{sec:results:manual}.

The manual validation (Section~\ref{sec:results:manual}) supports this combined
format: the text and the rendered path compensate for what the other cannot
show, so removing either lowers the fraction of checklist items the judge marks
as conveyed by the page, whereas the
per-step angle selection contributes little once text and a partially visible
motion are present. The validation also shows a marked gain when the judge is
briefed on the layout, suggesting that a short legend or worked example shipped
with the manual would help first-time users.

\subsection{Feedback Generation}
\label{sec:methods:feedback}
The pipeline produces feedback in two regimes. If the input fails one of the
checks in Sections~\ref{sec:methods:preprocessing}
or~\ref{sec:methods:feasibility}, the feedback explains why the assembly cannot
be built and, where possible, suggests a concrete fix. If the input is
buildable, the feedback instead proposes design changes that would make it
easier to assemble. In both regimes, the upstream stages do not hand the
feedback stage free text but a structured record for every detected problem: the
parts involved, the failure mode, the directions, displacements, or candidate
tools associated with it, and a rendered image highlighting the affected region.
This record is what is passed to a VLM for interpretation. 

\subsubsection{Feedback for unassemblable assemblies}
Four failure modes are distinguished. A single input can trigger several of them
at once, in which case all are reported. For each, the VLM receives the record
and its render and explains in two or three plain sentences what goes wrong and
which change to a part's shape would fix it; the prompt is given in
Appendix~\ref{sec:appendix:feedback_prompt}.

\begin{itemize}
  \item \textbf{No feasible sequence.} The disassembly search exhausts the
    candidate tree without ever finding a removable part. The remaining set must
    contain an interlocking group of parts that cannot be separated from one
    another; the set is rendered together with the corresponding part
    identifiers.
  \item \textbf{No stable subassembly.} Every candidate next removal leaves a
    configuration in which one or more parts fall. The falling parts are marked
    in the rendered output and their identifiers forwarded. When the initial
    assembly itself is unstable, the iterative joining procedure of
    Section~\ref{sec:methods:preprocessing} also returns a candidate set of
    fastening relations that would stabilise it, which is included as a concrete
    suggestion.
  \item \textbf{Tool inaccessible.} This mode is reported only once the search
    has been exhausted, not at the first failed tool check: no feasible sequence
    exists, and at least one remaining part could be removed with the rest of the
    assembly staying stable, but its required tool cannot be applied, either
    because the tool collides with surrounding geometry
    at the contact point, or because no path exists along which the tool can
    reach that contact point. In the collision case the obstructing parts are
    reported together with a render of the collision. In the path-blocked case
    the partial tool path is rendered up to its failure point.
  \item \textbf{Parts overlap.} Two or more parts overlap by more than the
    simulator can tolerate, as detected in preprocessing
    (Section~\ref{sec:methods:preprocessing}). The overlapping parts are flagged
    and the overlap region is rendered; since the pipeline does not move parts
    to remove the overlap, the fix is proposed by the VLM, as in
    Figure~\ref{fig:results:feedback_collision}.
\end{itemize}

\subsubsection{Feedback on assemblability improvements}
When the input is buildable, the pipeline produces a second kind of feedback
aimed at making the next revision of the design easier to assemble. The same
per-step information used by the cost function in Section~\ref{sec:methods:cost}
(reorientations, distance between steps, parts held in place, tool requirements)
is summarised for each step in the found sequence and passed to a VLM, one step
at a time, with a prompt that lists five common assembly
pitfalls~\cite{boothroyd2010product, butt2020}: poor clearance and access,
missing alignment features, awkward handling, ambiguous orientation, and parts
that fall before they are fastened (Appendix~\ref{sec:appendix:feedback_prompt}).
The VLM is asked to give feedback only where a step warrants a specific change,
and to return nothing otherwise, so that the suggestions stay actionable. Further
pitfalls can be added to the list; minimising the number of distinct fastener
types and lengths~\cite{boothroyd2010product}, for instance, is not yet included
and is left to future work.

\begin{figure*}[ht]
  \centering
  \includegraphics[width=\linewidth]{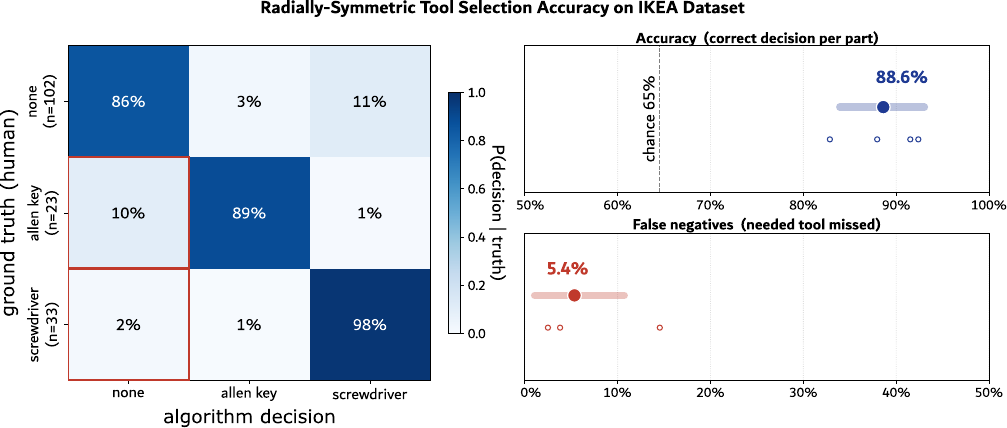}
  \caption{Tool decision accuracy on 158 parts from the IKEA 3D Assembly
Dataset~\cite{ikea3dassembly}, each added in its own assembly step. Of these, 56 require one of the two available tools. Each assembly is run five
  times, giving five decisions per part, which are averaged; 89\% of parts give the same answer
  across all five runs. Left: the row-normalised confusion matrix between the
  manual ground truth (rows) and the model decision (columns) over the three
  classes none, allen key, and screwdriver. The diagonal is the
  fraction of correct decisions per part, and the outlined cells are false
  negatives, parts that need a tool but are labelled as needing none. Right:
  two horizontal number lines giving the overall accuracy and the false-negative
  rate, each with a 95\% confidence interval, a chance reference, and one hollow dot per assembly.
  The model decides
  correctly on 88.6\% of parts (95\% CI 83.9 to 93.0), against a majority-class
  chance level of 64.6\% and a class-balanced accuracy of 90.7\%, and misses a
  needed tool on 5.4\% of the 56 tool parts (95\% CI 1.1 to 10.7), about 3
  parts.}
  \Description{Two-panel figure. Left: a three-by-three confusion matrix shaded
  from white to dark blue, with ground-truth classes none, allen key and
  screwdriver as rows and the algorithm's decision as columns. The diagonal
  reads 86 percent, 89 percent and 98 percent; the two cells in which a part
  needing a tool is labelled as needing none are outlined in red. Right: two
  horizontal number lines. The upper one plots accuracy at 88.6 percent with a
  confidence band, well to the right of a dashed chance reference at 65 percent;
  the lower one plots the false-negative rate at 5.4 percent near the left edge
  of a scale running to 50 percent. Hollow circles mark the individual
  assemblies on both lines.}
  \label{fig:results:tool_decisions}
\end{figure*}

\subsection{Experimental Evaluation}
\label{sec:methods:experimental_evaluation}

All language and vision tasks use OpenAI's multimodal GPT-5.4, except for the
design feedback shown in Section~\ref{sec:results:feedback}, which is generated
with the newer GPT-6 Sol. Physical simulation uses the differentiable, CPU-based
simulator of \citet{Wang2019}, parallelised across many CPU cores to
keep total simulation time tractable.

Two datasets are used. The IKEA 3D Assembly Dataset~\cite{ikea3dassembly, ikea2021} provides
realistic multi-part assemblies with the tool requirements and part diversity
needed to exercise the tool-application and sequencing stages; of its five
assemblies, only one meets the rigid-body assumption of the simulator, and the
other four are used to develop and evaluate the tool-application method
(Section~\ref{sec:methods:tools}). The second dataset, which we refer to as the
ASAP dataset, was released by \citet{tian2024asap}. Its 2{,}146 assemblies of
3 to more than 50 parts come from the Fusion 360 Gallery Assembly
Dataset~\cite{willis2021} of user-designed CAD models and from the mechanical
assemblies of \citet{lupinetti2019similarity}. Tian et al.\ kept only assemblies
with watertight part meshes that can be taken apart without deforming a part,
and removed parts that intersect others beyond a set distance. For the
240-assembly test split, they further dropped assemblies that were potentially
unstable in simulation and sampled up to ten at random per part count; the
remaining 1{,}906 form the training split. We use the test split for the timing
evaluation (Section~\ref{sec:results:sequence}), excluding six assemblies of 8
to 13 parts on which the weights are optimised, and for the runtime study
(Appendix~\ref{sec:appendix:runtime}), topped up from the training split where
it has too few assemblies. The manual evaluation uses ten hand-picked test
assemblies of 6 to 21 parts with easily recognised parts and function, plus the
qualifying IKEA assembly (Section~\ref{sec:results:manual}).

The tool-selection step is evaluated on the IKEA dataset, excluding one model
whose original assembly requires a tool not present in the catalogue. This leaves
four assemblies and 158 parts, of which 56 require one of the two available
tools, a Phillips screwdriver and a hex key (chosen to match the tools shipped
with the actual products). To reduce run-to-run variability in the VLM, each 
assembly is run five times, giving five decisions per part, which are
averaged. Since every part is added in its own step, the 158 parts correspond to
158 assembly steps.

For the sequencing evaluation, we use every assembly of the test split with up
to 30 parts, 200 after excluding the training assemblies; the test split
contains none with fewer than 5 parts. Every method is run with beam width $k = 4$, a budget of
6{,}000 candidate checks and a limit of 12\,h per plan. Of the 200 assemblies,
47 are excluded because no stable initial pose exists with at most two parts
held, and 136 are planned completely by every method. Of the rest, one reaches
the search budget with the optimised DfA cost only, one exceeds the time limit for
the random sequence, the baseline and the DfA cost with reference weights, and the others fail for
every method. Tool checking is disabled for all sequencing
and runtime experiments, since the ASAP assemblies contain almost no
fasteners.

The upper limit of 30 parts has two reasons. The data thin out at larger part
counts: the test split holds at most ten assemblies per part count, and the
share of assemblies without a stable initial pose or without a complete
sequence grows with every added part. At the same time, the planning runtime
grows faster than linearly with the number of parts
(Appendix~\ref{sec:appendix:runtime}), reaching a mean of 2{,}560\,s per
assembly at 21 to 25 parts with single runs close to two hours, so every method
compared multiplies this cost. Section~\ref{sec:results:sequence} shows that the
relative gain does not change with assembly size within this range, which gives
no reason to expect a different result above it.

The manual is evaluated with MLLMs acting as evaluators, either as zero-shot
classifiers or as VLM-as-judge graders~\cite{gu2025surveyllmasajudge}. Several
studies~\cite{han2025judgesverdictcomprehensiveanalysis,
chen2024mllmasajudgeassessingmultimodalllmasajudge,
lee2024prometheusvisionvisionlanguagemodeljudge} show that, particularly on a
comparative rather than absolute judgement task, VLMs are a reliable proxy for
human assessment, which underwrites the evaluation methodology of
Section~\ref{sec:results}.

\section{Results}
\label{sec:results}

We evaluate each of four pipeline
components in turn: the tool-application stage against manually
labelled tools and contact axes (Section~\ref{sec:results:tool}); the sequence
planner, with and without subassembly detection, against the 
assembly-time proxy, relative to an outermost-part-first baseline taken from
\citet{tian2024asap} (Section~\ref{sec:results:sequence}); the generated manuals through a
VLM-as-judge protocol (Section~\ref{sec:results:manual}); and the design feedback
qualitatively on representative failure cases (Section~\ref{sec:results:feedback}).

\subsection{Tool Application}
\label{sec:results:tool}
The tool-application stage makes two decisions that can be checked against
ground-truth labels (Section~\ref{sec:methods:tools}): which tool, if any, is
required to install a given part, and where the chosen tool contacts the part
and along which direction.

\subsubsection{Tool Decision}
As shown in Figure~\ref{fig:results:tool_decisions}, the VLM identifies the correct tool for 88.6\% of the 158 parts in the IKEA 3D Assembly
Dataset~\cite{ikea3dassembly}. It recognises that a tool is
required in 94.6\% of the parts that need one, with a false-negative rate of 5.4\%;
the remaining errors are false positives, where the model selects a tool for a
part that does not need one. Accuracy on the two tools individually is 97.6\%
for the Phillips screwdriver and 88.7\% for the hex key. Most errors fall on
visually ambiguous parts.

\subsubsection{Contact Decision}
On a catalogue of 10 screwing tools, the VLM is internally consistent across
repeated runs and selects both the correct application axis and direction for 5
of the 10 tools. The setup, per-tool breakdown, and error analysis are given in
Appendix~\ref{sec:appendix:contact}.

\subsection{Sequence Planning}
\label{sec:results:sequence}

\subsubsection{Sequence and subassembly validation}
We compare four sequence planners to a \emph{baseline} from \citet{tian2024asap} that prefers parts whose centre of mass lies
furthest from the centre of the assembly's bounding box. All planners use the same feasibility checks, stability checks, and beam
width ($k = 4$). \emph{Random sequence} draws the next beam at random from the feasible children
of the current one (a single seed). \emph{DfA cost} and \emph{DfA cost optimised} rank
partial sequences by the DfA cost $J$, with the reference weights of
Table~\ref{tab:methods:cost_functions} and with the weights optimised as in
Section~\ref{sec:methods:weight_optim}, respectively. \emph{DfA cost optimised +
subassemblies} adds the
subassembly plan of Section~\ref{sec:methods:subassembly} to the optimised sequence
and carries it out wherever a plan is found. Its time is predicted for one
worker, who builds the subassemblies one after the other, joining steps
included, and for two workers, who build the two halves of the outermost split
in parallel; the two-worker time is the makespan.

\begin{figure*}[t]
  \centering
  \includegraphics[width=\linewidth]{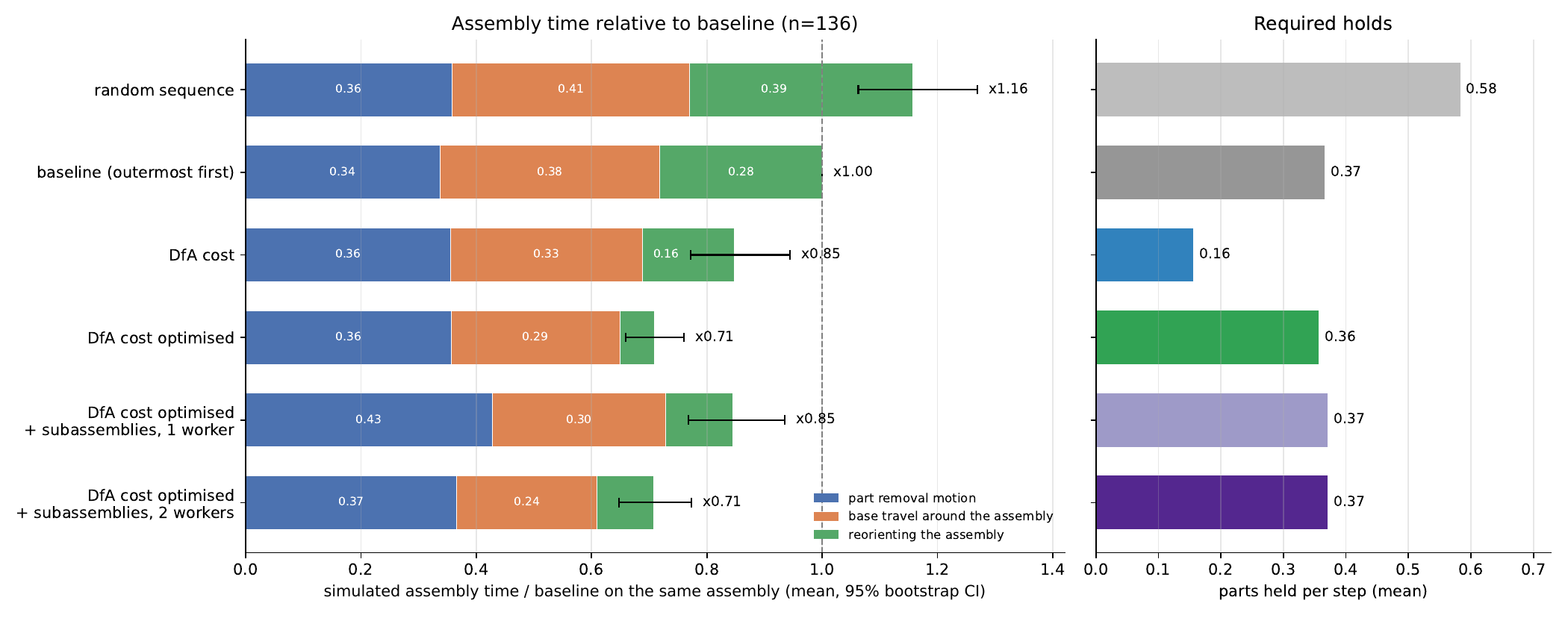}
  \caption{Simulated assembly time (Section~\ref{sec:methods:weight_optim}) on
  136 assemblies of 5 to 30 parts relative to the
  outermost-part-first baseline of \citet{tian2024asap} on the same assembly
  (dashed line at 1.0). Left: the time is split into part removal motion, base
  travel around the assembly, and reorientation of the assembly. Each component
  is divided by the baseline's total on the same assembly and averaged
  arithmetically, so the segments add up to the bar; the error bar is the 95\%
  bootstrap confidence interval of the total. For two workers the bar is the
  makespan, split into the same three components. These arithmetic
  means are higher than the geometric means of Table~\ref{tab:results:timing}.
  Right: mean number of parts that must be held per step, which the proxy does
  not charge for. The optimised DfA cost gives the lowest time on a single
  worker, mainly through less reorientation and base travel, while the DfA cost
  with reference weights holds the fewest parts.}
  \Description{Two-panel horizontal bar chart with six rows: random sequence,
  baseline (outermost first), DfA cost, DfA cost optimised, DfA cost optimised plus
  subassemblies with one worker, and DfA cost optimised plus subassemblies with
  two workers. The left panel shows predicted assembly time relative to the
  baseline, as stacked bars of part removal motion, base travel and
  reorientation, with a dashed line at 1.0. The totals are 1.16, 1.00, 0.85,
  0.71, 0.85 and 0.71. Part motion stays between 0.34 and 0.37 except for
  subassemblies with one worker (0.43); base travel falls from 0.41 for the
  random sequence to 0.29 for the optimised DfA cost and 0.24 for two workers;
  reorientation falls from 0.39 for the random sequence and 0.28 for the
  baseline to a thin segment for the optimised DfA cost. The right panel shows
  parts held per step: 0.58 for the random sequence, 0.37 for the baseline,
  0.16 for the DfA cost, 0.36 for the optimised DfA cost and 0.37 for both
  subassembly variants.}
  \label{fig:results:sequence_timing}
\end{figure*}

\begin{table}[t]
  \centering
  \caption{Predicted assembly time relative to the baseline: geometric mean of the
  per-assembly ratios with its 95\% bootstrap confidence interval, and the number
  of assemblies on which a method is faster than the baseline.}
  \label{tab:results:timing}
  \begin{tabular}{@{}lcc@{}}
    \toprule
    \textbf{Method} & \textbf{Time vs.\ baseline} & \textbf{Faster} \\
    \midrule
    Random sequence & $\times$1.05 [0.98, 1.13] & 62/136 \\
    DfA cost & $\times$0.76 [0.70, 0.82] & 104/136 \\
    DfA cost optimised & $\times$0.65 [0.61, 0.70] & 120/136 \\
    \quad + sub., 1 worker & $\times$0.75 [0.69, 0.81] & 109/136 \\
    \quad + sub., 2 workers & $\times$0.63 [0.58, 0.68] & 116/136 \\
    \bottomrule
  \end{tabular}
\end{table}

Table~\ref{tab:results:timing} and Figure~\ref{fig:results:sequence_timing}
compare the methods on 136 assemblies.
With the optimised DfA cost, the planner reduces the predicted assembly time by
35\% relative to the baseline (geometric mean $\times$0.65, 95\% confidence interval
[0.61, 0.70]) and is faster on 120 of 136 assemblies. This relative gain is
roughly constant across assembly sizes, from $\times$0.62 at 5 to 9 parts to
$\times$0.66 at 25 to 30 parts, with no significant trend. The DfA cost with reference weights,
set by hand and never fitted to the proxy, reduces assembly time by 24\% ($\times$0.76).

The random control shows that the baseline is better than chance, above all in
the parts that must be held. The random sequence requires 0.58 held parts per step
against 0.37 for the baseline, significantly more, and is faster
than the baseline on only 62 of 136 assemblies. Its difference in time is not
significant. The arithmetic mean of Figure~\ref{fig:results:sequence_timing}
puts the random sequence at $\times$1.16, raised by a few assemblies on which it
is much slower; the geometric mean, which treats a speed-up and a slow-down by
the same factor alike, puts it at $\times$1.05 [0.98, 1.13]. The optimised
DfA cost is better than chance: $\times$0.62 relative to the random
sequence, and faster on 121 of 136 assemblies.

Part motion time is similar for all sequence planners: it takes 0.34 to 0.36 of the baseline's
total for every linear sequence (Figure~\ref{fig:results:sequence_timing}). The
gains come from less base travel, from 0.38 for the baseline to 0.29 for the
optimised DfA cost, and above all from less reorientation, where the optimised
DfA cost needs 0.06, about a fifth of the baseline's 0.28.

The two weight sets, for the DfA cost and DfA cost optimised planners, serve different goals. The reference weights for the DfA cost planner are chosen
for the manual and the person who follows it, while the optimised weights are
fitted to assembly time alone, within the limit of two held parts and the
physics checks that every step must pass. Relative to $w_{\text{hold}} = 1$, the
optimiser raises $w_{\text{pose}}$ from 0.67 to 51.1 and $w_{\text{dist}}$ from
0.33 to 41.8, raises $w_{\text{top}}$ only from 0.67 to 3.63, and lowers
$w_{\text{free}}$ from 0.2 to 0.072, so that blocked degrees of freedom no
longer influence which part is removed next
(Table~\ref{tab:methods:cost_functions}). The fitted weights thus favour few
reorientations and short moves between steps, the two components where the
optimised sequences save time, which shows that the five DfA terms can encode a
planning strategy for assembly time and not only one for a readable manual.

Due to different objective weights, DfA cost and DfA cost optimised planners also differ in mean parts held. The reference
weights (DfA cost) hold the fewest parts, 0.16 per step, less than half of any other
method and significantly fewer than the baseline. In contrast, the optimised weights primarily minimise the proxy
time, which has a lower cost on holds relative to the other DfA cost terms while keeping steps feasible (at most two parts held at once). 

A subassembly plan is found for 96 of the 136 assemblies. On 84 it is carried
out as planned; on the other 12 it fails the re-check in the simulator, and
these assemblies, like the 40 without a plan, use the linear sequence. The
outermost split usually comes early, at a median of 11\% into the sequence and
at the very first step in 35 of 96 plans. The halves are balanced (a median of 7
and 7 parts at 15 to 20 parts), and 27 of 96 plans split again inside a half.
Splitting into subassemblies does not by itself save time. With one worker, the
plans are significantly slower than the linear optimised sequence ($\times$1.15
[1.09, 1.20] relative to it; $\times$0.75 against $\times$0.65 relative to
the baseline): the joining steps add part motion (0.36 to 0.43 of the baseline's total)
and the halves need extra reorientation (0.06 to 0.12). What subassemblies
provide is parallel work. With two workers building the halves of the outermost
split in parallel, the makespan recovers this loss and is as fast as the linear
sequence ($\times$0.63 relative to the baseline; $\times$0.97 [0.92, 1.01] relative
to the linear sequence, not significant).

These assembly times in Figure~\ref{fig:results:sequence_timing} are found with the same proxy that is used to train the weights
(Section~\ref{sec:methods:weight_optim}), so part of the optimised weights'
advantage over the reference weights ($\times$0.86 [0.82, 0.91], faster on 82
of 136 assemblies) reflects the optimiser fitting the proxy.
The reference weights are not fitted to our approach for measuring assembly time and still reduce the time by 24\%.

Comparing a generated sequence with a manufacturer's order, such as those in
IKEA-Manual~\cite{wang2022ikeamanual}, is a natural external check, but
similarity to a reference order is a poor measure of quality: quite different
orders can be equally easy to follow, and symmetric parts allow several
equivalent orders. We therefore time all feasible orders permitted by the manual
of the step stool (Figure~\ref{fig:example}) with the same proxy. Our planner's
sequence (105.7\,s) falls within their range (84.3\,s to 146.8\,s), and the
faster official orders gain mainly from less base travel, because the manual
groups parts by region of the assembly.

As an informal check outside the proxy, one participant with no prior exposure
assembled the physical parts of a puzzle-like assembly three times: first without
a manual, to become familiar with the parts, then with the manual generated from
the baseline sequence, and finally with the one generated from the DfA-optimised
sequence. The baseline manual took roughly 50\% longer, mostly on steps that
joined parts that were far apart or followed no evident pattern; with a single
participant and a fixed order in which learning favours the last attempt, this is
anecdotal support rather than a measurement of assembly time.

\subsection{Assembly Manual}
\label{sec:results:manual}
\begin{figure}[t]
  \centering
  \includegraphics[width=\linewidth]{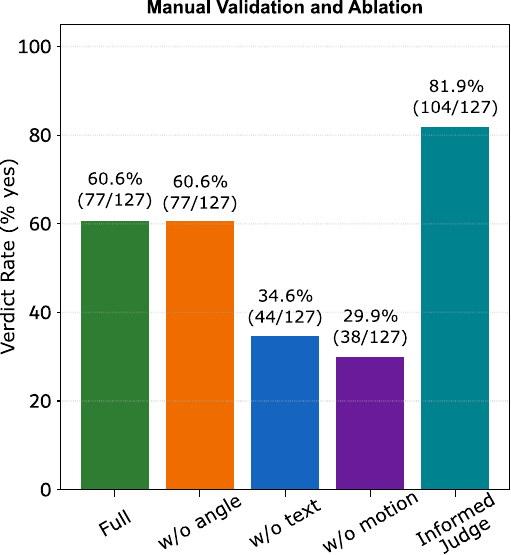}
  \caption{Manual validation with VLM judge on the 11-assembly test set. We compare the full manual (\textbf{Full} in green) with ablated versions excluding per-step camera-angle selection (\textbf{w/o angle} in orange), VLM-generated text instructions (\textbf{w/o text} in blue), and the rendered assembly path of the
    active part (\textbf{w/o motion} in purple). We also compare to a condition in which the judge VLM is given a short context document explaining how the
    manual layout is intended to be read (\textbf{Informed Judge} in turquoise).}
  \Description{A bar chart of the judge's verdict rate, in percent of items
  answered yes, for five conditions. The full manual and the fixed-angle variant
  are equal at 60.6 percent, 77 of 127 items. Removing the text drops the rate to
  34.6 percent, 44 of 127, and removing the path overlay drops it further to 29.9
  percent, 38 of 127. Giving the judge a pre-brief on the layout raises it to
  81.9 percent, 104 of 127, the tallest bar.}
  \label{fig:results:manual_validation}
\end{figure}

Figure~\ref{fig:results:manual_validation} presents the assessment of a VLM judge on the generated assembly manuals. 
The judge VLM is shown a generated manual page together with a checklist of items
the page should convey (the part being added, its direction of approach, any
required reorientation, and any required tool) and is asked, for each item,
whether the page alone contains enough information for a reader to determine it.
The output is a binary verdict per item. The judge does not carry out the step or
reconstruct the item itself; the verdicts therefore measure whether the
information is available on the page, not whether a reader would reconstruct it
correctly. The VLM-as-judge protocol has been shown to track human
assessments reliably when the judging task is binary and tightly
constrained~\cite{han2025judgesverdictcomprehensiveanalysis,chen2024mllmasajudgeassessingmultimodalllmasajudge,lee2024prometheusvisionvisionlanguagemodeljudge}.
The baseline manual generator produces, for each assembly step, a single page
that combines a coloured render of the action, a text instruction, and
conditional callouts for rotations and tools
(Section~\ref{sec:methods:manual}). A representative step is shown in
Figure~\ref{fig:appendix:manual_example} in the appendix. To isolate
the contribution of each design element, we ask the VLM-judge to also assess degraded variants of
the same manual:

\begin{itemize}
  \item \textbf{w/o angle} (orange):
    the per-step camera-angle selection of Section~\ref{sec:methods:manual} is
    skipped and every step is rendered from the same fixed angle.
  \item \textbf{w/o text} (blue): the VLM-generated text instructions are removed.
  \item \textbf{w/o motion} (purple): the rendered assembly path of the
    active part is removed from the page.
  \item \textbf{Informed Judge} (turquoise): on top of the full
    manual, the judge VLM is given a short context document explaining how the
    manual layout is intended to be read.
\end{itemize}

Across the 127 checklist items, the judge marks 77 (60.6\%) as conveyed by the
full manual. Among the degraded variants, removing the text and removing the path
overlay lower this to 44 (34.6\%) and 38 (29.9\%) items respectively, whereas
skipping the per-step angle selection leaves it unchanged. Supplying the judge with
the layout pre-brief raises it to 104 items (81.9\%).

\subsection{Design for Assembly Feedback}
\label{sec:results:feedback}

The feedback in both examples is generated with GPT-6 Sol
(Section~\ref{sec:methods:experimental_evaluation}), using the prompt in
Appendix~\ref{sec:appendix:feedback_prompt}.
Figure~\ref{fig:results:feedback_collision} shows an overlap between two parts
the pipeline has named the toothed drive shaft (red) and the long compression
spring (blue). The shaft is a tube with a crown of teeth at one end, and the
spring sits inside it, but the outer diameter of its coils is larger than the
hole through the shaft, so the coils cut into the inside wall along its whole
length. The overlap is shown in four views, with the two parts drawn translucent
so that the spring remains visible inside the shaft. The pipeline names the hole
through the shaft as the feature at fault and proposes to widen it just enough
for the spring to fit without touching the wall. The fix thus states which part
to change, which feature of that part, and by how much.

\begin{figure}[ht]
  \centering
  \includegraphics[width=\linewidth]{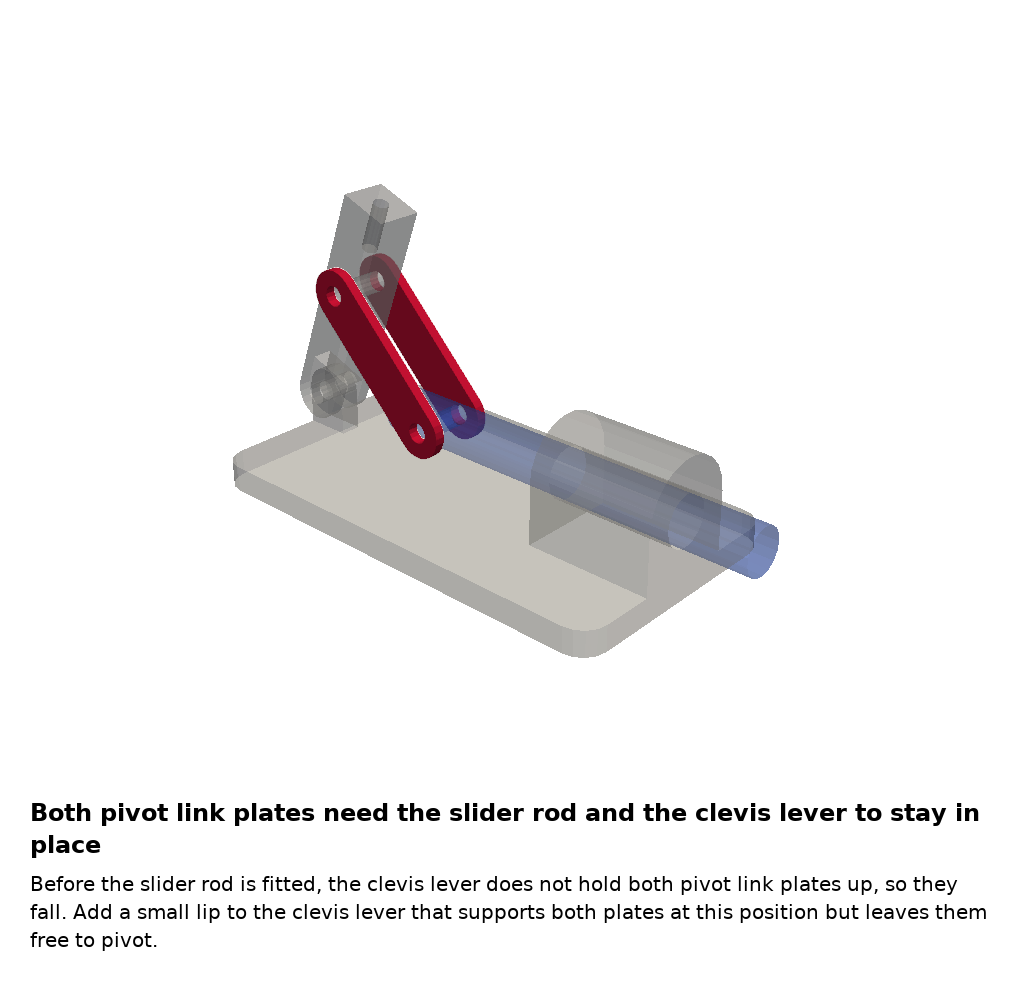}
  \caption{Instability case, as returned by the pipeline: a render of the partial
  assembly, followed by the generated title and suggestion. The two pivot link
  plates (red) are held up by the slider rod (blue) together with the clevis
  lever, the upright grey part on the left; before the slider rod is fitted, the plates fall. The suggested fix is a
  small lip on the clevis lever that supports both plates while leaving them free
  to pivot.}
  \Description{Translucent render of a small linkage mechanism on a flat base
  plate: an upright clevis lever on the left, a block on the right through which
  a blue slider rod runs, and two red link plates connecting the lever to the
  rod. The remaining parts are grey. Below the render, the pipeline's text reads:
  both pivot link plates need the slider rod and the clevis lever to stay in
  place; before the slider rod is fitted, the clevis lever does not hold the
  plates up, so they fall; add a small lip to the clevis lever that supports both
  plates but leaves them free to pivot.}
  \label{fig:results:feedback_stability}
\end{figure}

Figure~\ref{fig:results:feedback_stability} shows an assembly that fails the
stability check. Two pivot link plates (red) are held in place only once both the
slider rod (blue) and the clevis lever are fitted. At the step in which the
slider rod is not yet in place, the clevis lever alone cannot carry the plates,
and they fall. The pipeline identifies which parts the plates depend on and
proposes adding a small lip to the clevis lever that supports both plates at
this position while leaving them free to pivot, so that the plates stay put
before the slider rod is inserted without losing their function.

Measuring the end-to-end value of these suggestions would require a downstream
agent to act on them, which is left to future work, and is further limited by the
absence of any dataset pairing assemblies with labelled design defects or
matched defective and corrected versions; existing assemblability benchmarks
score whether a generated design can be built, not which change would fix
it~\cite{dong2026muse}. Within that constraint, the suggestions
nonetheless track the kinds of design issues a human reviewer would flag, a
necessary if not sufficient condition for them to serve as a corrective signal.

\begin{figure}[ht]
  \centering
  \includegraphics[width=\linewidth]{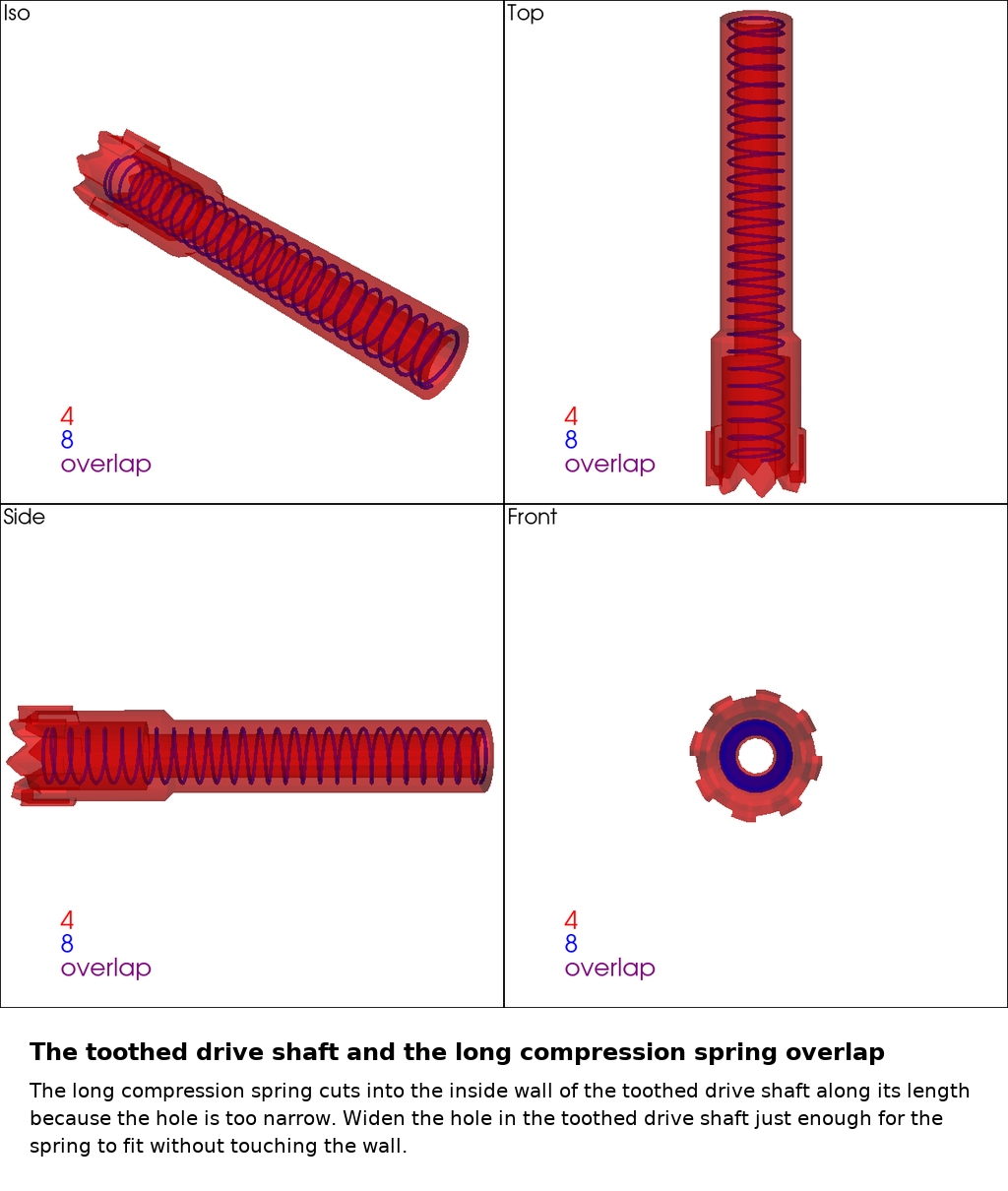}
  \caption{Overlap case, as returned by the pipeline: four views of the two
  overlapping parts, followed by the generated title and suggestion. A
  compression spring (blue, part 8) sits inside a toothed drive shaft (red,
  part 4), a tube with a crown of teeth at one end, but the spring's outer
  diameter is larger than the hole through the shaft, so the coils cut into the
  inside wall along its whole length. The suggested fix is to widen the hole just
  enough for the spring to fit without touching the wall.}
  \Description{Four translucent renders, labelled Iso, Top, Side and Front, of a
  red tubular drive shaft with a crown of teeth at one end and a blue coil
  spring running through it. In the top and side views the spring's coils reach
  into the inside wall over the full length of the shaft; the front view looks
  down the axis and shows the spring filling the hole. Below the renders, the
  pipeline's text reads: the toothed drive shaft and the long compression spring
  overlap; the spring cuts into the inside wall of the shaft along its length
  because the hole is too narrow; widen the hole in the shaft just enough for the
  spring to fit without touching the wall.}
  \label{fig:results:feedback_collision}
\end{figure}

\section{Limitations}
\label{sec:limitations}
The dominant runtime bottleneck of the pipeline is the underlying physics
simulation in the CPU-only simulator~\cite{Wang2019}, and the per-frontier
simulation cost dominates the search. Measured on ASAP assemblies of 5 to
26 parts (Appendix~\ref{sec:appendix:runtime}), sequence-planning runtime grows
faster than linearly with the number of parts, so the time spent per part rises
with assembly size. Most of this increase comes from the stability check, which
simulates every part still in the assembly at once and therefore becomes more
expensive as parts are added. Path finding, by contrast, stays roughly constant,
since the rest of the assembly can be treated as a single fixed body, leaving
only two bodies to simulate per candidate. Scaling to larger assemblies will
therefore require reducing the simulation cost per candidate, above all in the
stability check, for instance by caching the simulator's partial state across
candidates that differ by a single part. Path finding could be shortened further
by adding a fast geometric prefilter before the simulator. Ray casting or rasterisation on graphics hardware computes directional
blocking exactly and cheaply~\cite{spitz2000graphics, khardekar2006mold}, but only
for straight translations. In the simulator, a part pushed along a direction may
deviate from a straight line and rotate, for instance sliding along a contact or
following a slightly curved fit (Section~\ref{sec:methods:subassembly}), and small
interpenetrations are tolerated. These imperfect paths are much closer to how a
human or a robot actually inserts a part than an exact straight-line test, which
is the main reason the simulator is kept as the final judge. A prefilter should therefore only accept, never
reject: a candidate found free along a straight line by ray casting is free and
needs no simulation, whereas a candidate blocked in every straight direction must
still be passed to the simulator. The directional blocking matrix already used in
subassembly detection (Section~\ref{sec:methods:subassembly}) could serve the same
role for candidates it has already decided.

The tool-application stage is limited to radially symmetric tools and is fragile
overall. The contact-point and axis selection (Section~\ref{sec:methods:tools})
uses a convex-decomposition heuristic tailored primarily to screwing
tools; when the relevant application axis is not a principal axis of one of the
convex components, the heuristic does not surface a correct candidate and the VLM
has nothing correct to pick from. The applicability check also assumes that a
tool needs only a straight clear path to its contact point, which is a reasonable
model for screwdrivers but fails for tools like a hammer, where additional swing
clearance is required and is not currently accounted for; a per-tool use volume
in the sense of \citet{wilson1996tools} would capture this, at the cost of an
annotation per tool.

The simulator considers only rigid parts. This rigid-body assumption is a
fundamental restriction, excluding assemblies that rely on friction or compliance
to hold together, a class that our sequencing logic would otherwise be able to
handle. Snap-fit joints, although common in practice, fall under the same
restriction: inserting one requires the elastic deflection of a hook or rim, so
a rigid simulator finds the part blocked in both directions, and many snap-fits
cannot be released once engaged, so the joint cannot be reached by disassembly.
One path to supporting them is to detect snap-fit features, for instance with the
VLM that already names each part (Section~\ref{sec:methods:naming}), and to model
the flagged contacts as compliant in the simulator, or to allow a one-way
insertion step for them in the search.

Because sequencing is done step-by-step and subassemblies are only explored once
a sequence has been found, the current approach does not handle assemblies that
strictly require several parts to be removed simultaneously. The existing
subassembly algorithm (Section~\ref{sec:methods:subassembly}) could be repurposed
to guess candidate simultaneous removals, but validating them is intractable with
the current simulator, so we leave this to future work.

\section{Conclusions}
\label{sec:conclusion}
We present an end-to-end approach that turns a raw mesh assembly, with no joint
metadata, fastener annotations, or other structured input, into either a
step-by-step assembly manual or a structured failure report. Four components of a
manufacturing plan are produced autonomously: an assembly tool list, a DfA-aware
assembly sequence together with parallelisable subassemblies, a human-readable
manual, and design feedback for improving assemblability. Physical feasibility and
sequencing are handled by systematically disassembling the object in a physics
simulator under a cost function that encodes design-for-assembly principles, while
tool labelling, manual generation, and feedback rely on multimodal large language
models. DfA-aware sequence planning reduces the robot-arm proxy assembly time by
35\% on 136 assemblies of 5 to 30 parts. Splitting the sequence into
subassemblies does not shorten it further, but it allows two workers to build
the subassemblies in parallel; with two workers the reduction is 37\%, on par
with the linear sequence. The correct tool is
identified for 88.6\% of parts, and a vision-language-model judge,
checking which items of a per-step checklist each manual page conveys, finds that
the rendered motion and the generated text each carry a substantial share of that
information.

The contributions of this work bear directly on two ongoing trends in mechanical
design. The first is the rise of Text- and Image-to-CAD models such as
CAD-Llama~\cite{li2025cadllamaleveraginglargelanguage} and
ArtiCAD~\cite{shui2026articadarticulatedcadassembly}, which can generate
mechanical components and articulated assemblies directly from natural language
or image input, but whose spatial-reasoning limits leave the resulting geometry
unverified. For these, the pipeline can act as a deterministic verifier and
feedback channel that closes the loop toward fully autonomous CAD generation:
the upstream model generates a candidate assembly, and the pipeline returns
either a manual confirming buildability or a structured failure report that the
model can incorporate as a corrective signal, in the manner of agentic design
loops that place an explicit physical verifier between generation
steps~\cite{berger2026physicsinloop}. The second is the more traditional
design-for-production gap. Even for assemblies designed by hand, no widely
available lightweight tool today takes a finished CAD model and returns both a
feasible assembly plan and an instruction manual without manual annotation work.
The pipeline is light enough to fit into either setting and is, to our knowledge,
the first to do so end-to-end from raw mesh input.


\section{Ethics and Privacy Statement}
If this approach is used uncritically, particularly with load-bearing or otherwise dangerous assemblies, this work could present safety concerns. The
tool-decision stage fails to identify a required tool on 5.4\% of the parts that
need one (Section~\ref{sec:results:tool}). Generated manuals should be reviewed by a competent person before usage.

\bibliographystyle{ACM-Reference-Format}
\bibliography{references}

@book{boothroyd2010product,
  title={Product design for manufacture and assembly},
  author={Boothroyd, Geoffrey and Dewhurst, Peter and Knight, Winston A},
  year={2010},
  publisher={CRC Press},
  address={Boca Raton, FL, USA},
  edition={3rd}
}

@misc{gu2025surveyllmasajudge,
      title={A Survey on LLM-as-a-Judge}, 
      author={Jiawei Gu and Xuhui Jiang and Zhichao Shi and Hexiang Tan and Xuehao Zhai and Chengjin Xu and Wei Li and Yinghan Shen and Shengjie Ma and Honghao Liu and Saizhuo Wang and Kun Zhang and Yuanzhuo Wang and Wen Gao and Lionel Ni and Jian Guo},
      year={2025},
      eprint={2411.15594},
      archivePrefix={arXiv},
      primaryClass={cs.CL},
      url={https://arxiv.org/abs/2411.15594}, 
}

@misc{shui2026articadarticulatedcadassembly,
      title={ArtiCAD: Articulated CAD Assembly Design via Multi-Agent Code Generation}, 
      author={Yuan Shui and Yandong Guan and Zhanwei Zhang and Juncheng Hu and Jing Zhang and Dong Xu and Qian Yu},
      year={2026},
      eprint={2604.10992},
      archivePrefix={arXiv},
      primaryClass={cs.CV},
      url={https://arxiv.org/abs/2604.10992}, 
}

@misc{li2025cadllamaleveraginglargelanguage,
      title={CAD-Llama: Leveraging Large Language Models for Computer-Aided Design Parametric 3D Model Generation}, 
      author={Jiahao Li and Weijian Ma and Xueyang Li and Yunzhong Lou and Guichun Zhou and Xiangdong Zhou},
      year={2025},
      eprint={2505.04481},
      archivePrefix={arXiv},
      primaryClass={cs.CV},
      url={https://arxiv.org/abs/2505.04481}, 
}

@article{Khan2025,
	author = {Khan, Muhammad Tayyab and Chen, Lequn and Ng, Ye Han and Feng, Wenhe and Tan, Nicholas Yew Jin and Moon, Seung Ki},
	doi = {10.1115/1.4069266},
	issn = {1530-9827},
	journal = {Journal of Computing and Information Science in Engineering},
	month = {08},
	number = {10},
	pages = {104501},
	title = {Leveraging Vision-Language Models for Manufacturing Feature Recognition in Computer-Aided Designs},
	url = {https://doi.org/10.1115/1.4069266},
	volume = {25},
	year = {2025}
}

@article{doris2024designqa,
      title   = {{DesignQA}: A Multimodal Benchmark for Evaluating Large Language Models' Understanding of Engineering Documentation},
      author  = {Doris, Anna C. and Grandi, Daniele and Tomich, Ryan and Alam, Md Ferdous and Ataei, Mohammadmehdi and Cheong, Hyunmin and Ahmed, Faez},
      journal = {Journal of Computing and Information Science in Engineering},
      volume  = {25},
      number  = {2},
      pages   = {021009},
      year    = {2025},
      publisher = {ASME}
}

@article{tian2022assemble,
  title={Assemble Them All: Physics-Based Planning for Generalizable Assembly by Disassembly},
  author={Tian, Yunsheng and Xu, Jie and Li, Yichen and Luo, Jieliang and Sueda, Shinjiro and Li, Hui and Willis, Karl D.D. and Matusik, Wojciech},
  journal={ACM Trans. Graph.},
  volume={41},
  number={6},
  articleno={278},
  numpages={15},
  year={2022},
  publisher={ACM}
}

@inproceedings{tian2024asap,
  title={{ASAP}: Automated sequence planning for complex robotic assembly with physical feasibility},
  author={Tian, Yunsheng and Willis, Karl DD and Al Omari, Bassel and Luo, Jieliang and Ma, Pingchuan and Li, Yichen and Javid, Farhad and Gu, Edward and Jacob, Joshua and Sueda, Shinjiro and others},
  booktitle={2024 IEEE International Conference on Robotics and Automation (ICRA)},
  pages={4380--4386},
  year={2024},
  publisher={IEEE},
  address={Yokohama, Japan}
}

@misc{hartmann2025unifiedframeworkautomatedassembly,
      title={A Unified Framework for Automated Assembly Sequence and Production Line Planning using Graph-based Optimization}, 
      author={Christoph Hartmann and Marios Demetriades and Kevin Prüfer and Zichen Zhang and Klaus Spindler and Stefan Weltge},
      year={2025},
      eprint={2512.13219},
      archivePrefix={arXiv},
      primaryClass={cs.RO},
      url={https://arxiv.org/abs/2512.13219}, 
}

@article{Dini1992,
    title = {Automated Sequencing and Subassembly Detection in Assembly Planning},
    journal = {CIRP Annals},
    volume = {41},
    number = {1},
    pages = {1-4},
    year = {1992},
    issn = {0007-8506},
    doi = {10.1016/S0007-8506(07)61140-8},
    url = {https://www.sciencedirect.com/science/article/pii/S0007850607611408},
    author = {G. Dini and M. Santochi}
}

@article{gao2025,
	author = {Gao, Z. H. and Yu, Fei and Cao, Jie and Yang, Xin and Xie, J. H. and Guan, Z. Q.},
	date = {2025/07/02},
	doi = {10.1038/s41598-025-04160-8},
	id = {Gao2025},
	isbn = {2045-2322},
	journal = {Scientific Reports},
	number = {1},
	pages = {23153},
	title = {Robust mesh generation for electromagnetic models with geometric defects through node alignment and mesh boolean operations},
	url = {https://doi.org/10.1038/s41598-025-04160-8},
	volume = {15},
	year = {2025}
}

@Article{butt2020,
    AUTHOR = {Butt, Javaid and Jedi, Sunny},
    TITLE = {Redesign of an In-Market Conveyor System for Manufacturing Cost Reduction and Design Efficiency Using DFMA Methodology},
    JOURNAL = {Designs},
    VOLUME = {4},
    YEAR = {2020},
    NUMBER = {1},
    articleno = {6},
    URL = {https://www.mdpi.com/2411-9660/4/1/6},
    ISSN = {2411-9660},
    DOI = {10.3390/designs4010006}
}

@mastersthesis{jiang1994optimization,
  author    = {Jiang, J.},
  title     = {Optimization of Robotic Mechanical Assembly Sequence Utilizing Probabilistic Constraints},
  school    = {West Virginia University},
  year      = {1994},
  doi       = {10.33915/etd.9111},
  url       = {https://doi.org/10.33915/etd.9111}
}

@article{bahuba2019,
	author = {Bahubalendruni, M. V. A. Raju and Gulivindala, Anil Kumar and Varupala, S. S. V. Prasad and Palavalasa, Dileep Kumar},
	date = {2019/07/08},
	doi = {10.1007/s12046-019-1157-2},
	id = {Bahubalendruni2019},
	isbn = {0973-7677},
	journal = {S{\=a}dhan{\=a}},
	number = {8},
	pages = {174},
	title = {Optimal Assembly Sequence generation through computational approach},
	url = {https://doi.org/10.1007/s12046-019-1157-2},
	volume = {44},
	year = {2019}}

@ARTICLE{watson2019,
  author={Watson, James and Hermans, Tucker},
  journal={IEEE Robotics and Automation Letters}, 
  title={Assembly Planning by Subassembly Decomposition Using Blocking Reduction}, 
  year={2019},
  volume={4},
  number={4},
  pages={4054-4061},
  doi={10.1109/LRA.2019.2929995}}

@article{wilson1994,
    title = {Geometric reasoning about mechanical assembly},
    journal = {Artificial Intelligence},
    volume = {71},
    number = {2},
    pages = {371-396},
    year = {1994},
    issn = {0004-3702},
    doi = {10.1016/0004-3702(94)90048-5},
    url = {https://www.sciencedirect.com/science/article/pii/0004370294900485},
    author = {Randall H. Wilson and Jean-Claude Latombe}
}

@article{gulivindala2020,
    author = {Gulivindala, Anil Kumar and Bahubalendruni, M.V.A. Raju and Varupala, S.S. Vara Prasad and K., Sankaranarayanasamy},
    title = {A heuristic method with a novel stability concept to perform parallel assembly sequence planning by subassembly detection},
    journal = {Assembly Automation},
    volume = {40},
    number = {5},
    pages = {779-787},
    year = {2020},
    month = {08},
    issn = {0144-5154},
    doi = {10.1108/AA-01-2020-0017},
    url = {https://doi.org/10.1108/AA-01-2020-0017},
}

@article{demello1990,
      author = {Sanderson, Arthur C. and de Mello, Luiz S. Homem and Zhang, Hui},
      title = {Assembly Sequence Planning},
      journal = {AI Magazine},
      volume = {11},
      number = {1},
      pages = {62-81},
      doi = {10.1609/aimag.v11i1.824},
      year = {1990}
}

@misc{han2025judgesverdictcomprehensiveanalysis,
      title={Judge's Verdict: A Comprehensive Analysis of LLM Judge Capability Through Human Agreement}, 
      author={Steve Han and Gilberto Titericz Junior and Tom Balough and Wenfei Zhou},
      year={2025},
      eprint={2510.09738},
      archivePrefix={arXiv},
      primaryClass={cs.CL},
      url={https://arxiv.org/abs/2510.09738}, 
}

@inproceedings{chen2024mllmasajudgeassessingmultimodalllmasajudge,
      title     = {{MLLM-as-a-Judge}: Assessing Multimodal {LLM-as-a-Judge} with Vision-Language Benchmark},
      author    = {Chen, Dongping and Chen, Ruoxi and Zhang, Shilin and Wang, Yaochen and Liu, Yinuo and Zhou, Huichi and Zhang, Qihui and Wan, Yao and Zhou, Pan and Sun, Lichao},
      booktitle = {Proceedings of the 41st International Conference on Machine Learning (ICML)},
      series    = {Proceedings of Machine Learning Research},
      volume    = {235},
      pages     = {6562--6595},
      year      = {2024},
      publisher = {PMLR},
      address   = {Vienna, Austria}
}

@inproceedings{lee2024prometheusvisionvisionlanguagemodeljudge,
      title     = {{Prometheus-Vision}: Vision-Language Model as a Judge for Fine-Grained Evaluation},
      author    = {Lee, Seongyun and Kim, Seungone and Park, Sue Hyun and Kim, Geewook and Seo, Minjoon},
      booktitle = {Findings of the Association for Computational Linguistics: ACL 2024},
      pages     = {11286--11315},
      year      = {2024},
      publisher = {Association for Computational Linguistics},
      doi       = {10.18653/v1/2024.findings-acl.672},
      address   = {Bangkok, Thailand}
}

@article{Wang2019,
  author = {Wang, Ying and Weidner, Nicholas J. and Baxter, Margaret A. and Hwang, Yura and Kaufman, Danny M. and Sueda, Shinjiro},
  title = {\textsc{RedMax}: Efficient \& Flexible Approach for Articulated Dynamics},
  year = {2019},
  issue_date = {July 2019},
  publisher = {ACM},
  address = {New York, NY, USA},
  volume = {38},
  number = {4},
  issn = {0730-0301},
  url = {https://doi.org/10.1145/3306346.3322952},
  doi = {10.1145/3306346.3322952},
  journal = {{ACM} Trans.\ Graph.},
  month = jul,
  articleno = {104},
  numpages = {10}
}

@article{michas2000,
      author = {Michas, Irene C and Berry, Dianne C},
      title = {Learning a procedural task: effectiveness of multimedia presentations},
      journal = {Applied Cognitive Psychology},
      volume = {14},
      number = {6},
      pages = {555-575},
      doi = {10.1002/1099-0720(200011/12)14:6<555::AID-ACP677>3.0.CO;2-4},
      year = {2000}
}

@article{boothroyd1994,
  author    = {Boothroyd, Geoffrey},
  title     = {Product design for manufacture and assembly},
  journal   = {Computer-Aided Design},
  volume    = {26},
  number    = {7},
  pages     = {505--520},
  year      = {1994},
  publisher = {Elsevier}
}

@article{fitts1954,
  author    = {Fitts, Paul M.},
  title     = {The information capacity of the human motor system in
               controlling the amplitude of movement},
  journal   = {Journal of Experimental Psychology},
  volume    = {47},
  number    = {6},
  pages     = {381--391},
  year      = {1954},
  doi       = {10.1037/h0055392}
}

@article{binder2025subgoals,
  author  = {Binder, Felix J. and Mattar, Marcelo G. and Kirsh, David J. and Fan, Judith E.},
  title   = {Humans Select Subgoals That Balance Immediate and Future Cognitive Costs During Physical Assembly},
  journal = {Cognitive Science},
  volume  = {49},
  number  = {11},
  pages   = {e70135},
  year    = {2025},
  doi     = {10.1111/cogs.70135}
}

@article{picard2025concept,
  author    = {Picard, Cyril and Edwards, Kristen M. and Doris, Anna C. and
               Man, Brandon and Giannone, Giorgio and Alam, Md Ferdous and
               Ahmed, Faez},
  title     = {From concept to manufacturing: evaluating vision-language
               models for engineering design},
  journal   = {Artificial Intelligence Review},
  volume    = {58},
  number    = {9},
  pages     = {288},
  year      = {2025},
  doi       = {10.1007/s10462-025-11290-y},
  publisher = {Springer}
}

@misc{dong2026muse,
  title         = {MUSE: Benchmarking Manufacturable, Functional, and Assemblable Text-to-CAD Generation},
  author        = {Dong, Xiaoyu and Li, Zhi and Wu, Xiao-Ming},
  year          = {2026},
  eprint        = {2605.28579},
  archivePrefix = {arXiv},
  primaryClass  = {cs.CV},
  url           = {https://arxiv.org/abs/2605.28579}
}

@inproceedings{akiba2019optuna,
  title     = {Optuna: A Next-generation Hyperparameter Optimization Framework},
  author    = {Akiba, Takuya and Sano, Shotaro and Yanase, Toshihiko and Ohta, Takeru and Koyama, Masanori},
  booktitle = {Proceedings of the 25th ACM SIGKDD International Conference on Knowledge Discovery \& Data Mining},
  pages     = {2623--2631},
  year      = {2019},
  publisher = {ACM},
  doi       = {10.1145/3292500.3330701},
  address   = {Anchorage, AK, USA}
}

@inproceedings{bergstra2011tpe,
  title     = {Algorithms for Hyper-Parameter Optimization},
  author    = {Bergstra, James and Bardenet, R{\'e}mi and Bengio, Yoshua and K{\'e}gl, Bal{\'a}zs},
  booktitle = {Advances in Neural Information Processing Systems (NeurIPS)},
  volume    = {24},
  year      = {2011},
  pages     = {2546--2554},
  publisher = {Curran Associates, Inc.},
  address   = {Granada, Spain}
}

@misc{ikea3dassembly,
  author       = {{Inter IKEA Systems B.V.}},
  title        = {The {IKEA} {3D} Assembly Dataset},
  year         = {2021},
  howpublished = {\url{https://github.com/IKEA/IKEA3DAssemblyDataset}},
  note         = {Licensed under CC BY-NC-SA 4.0}
}

@misc{ikea2021,
       title={IKEA Object State Dataset: A 6DoF object pose estimation dataset and benchmark for multi-state assembly objects},
       author={Yongzhi Su and Mingxin Liu and Jason Rambach and Antonia Pehrson and Anton Berg and Didier Stricker},
       year={2021},
       eprint={2111.08614},
       archivePrefix={arXiv},
       primaryClass={cs.CV}
}

@article{lupinetti2019similarity,
  title={Content-based multi-criteria similarity assessment of {CAD} assembly models},
  author={Lupinetti, Katia and Giannini, Franca and Monti, Marina and Pernot, Jean-Philippe},
  journal={Computers in Industry},
  volume={112},
  pages={103111},
  year={2019},
  doi={10.1016/j.compind.2019.07.001}
}

@inproceedings{willis2021,
  title={JoinABLe: Learning Bottom-up Assembly of Parametric CAD Joints},
  author={Willis, Karl DD and Jayaraman, Pradeep Kumar and Chu, Hang and Tian, Yunsheng and Li, Yifei and Grandi, Daniele and Sanghi, Aditya and Tran, Linh and Lambourne, Joseph G and Solar-Lezama, Armando and Matusik, Wojciech},
  booktitle={Proceedings of the IEEE/CVF Conference on Computer Vision and Pattern Recognition (CVPR)},
  pages={15828--15839},
  year={2022},
  doi={10.1109/CVPR52688.2022.01539},
  publisher = {IEEE},
  address   = {New Orleans, LA, USA}
}

@article{defazio1987,
  author  = {De Fazio, Thomas L. and Whitney, Daniel E.},
  title   = {Simplified generation of all mechanical assembly sequences},
  journal = {IEEE Journal on Robotics and Automation},
  volume  = {3},
  number  = {6},
  pages   = {640--658},
  year    = {1987},
  doi     = {10.1109/JRA.1987.1087132}
}

@article{demello1990andor,
  author  = {Homem de Mello, Luiz S. and Sanderson, Arthur C.},
  title   = {{AND/OR} graph representation of assembly plans},
  journal = {IEEE Transactions on Robotics and Automation},
  volume  = {6},
  number  = {2},
  pages   = {188--199},
  year    = {1990},
  doi     = {10.1109/70.54734}
}

@article{jimenez2013survey,
  author  = {Jim{\'e}nez, Pablo},
  title   = {Survey on assembly sequencing: a combinatorial and geometrical perspective},
  journal = {Journal of Intelligent Manufacturing},
  volume  = {24},
  number  = {2},
  pages   = {235--250},
  year    = {2013},
  doi     = {10.1007/s10845-011-0578-5}
}

@article{ghandi2015review,
  author  = {Ghandi, Sepehr and Masehian, Ellips},
  title   = {Review and taxonomies of assembly and disassembly path planning problems and approaches},
  journal = {Computer-Aided Design},
  volume  = {67--68},
  pages   = {58--86},
  year    = {2015},
  doi     = {10.1016/j.cad.2015.05.001}
}

@article{lambert2003disassembly,
  author  = {Lambert, A. J. D.},
  title   = {Disassembly sequencing: a survey},
  journal = {International Journal of Production Research},
  volume  = {41},
  number  = {16},
  pages   = {3721--3759},
  year    = {2003},
  doi     = {10.1080/0020754031000120078}
}

@inproceedings{huang2021newanalog,
  author    = {Huang, Yijiang and Leung, Pok Yin Victor and Garrett, Caelan Reed and Gramazio, Fabio and Kohler, Matthias and Mueller, Caitlin},
  title     = {The New Analog: A Protocol for Linking Design and Construction Intent with Algorithmic Planning for Robotic Assembly of Complex Structures},
  booktitle = {Proceedings of the 6th Annual ACM Symposium on Computational Fabrication (SCF)},
  articleno = {9},
  numpages  = {17},
  year      = {2021},
  publisher = {ACM},
  doi       = {10.1145/3485114.3485122},
  address   = {Virtual Event, USA}
}

@inproceedings{vallat2023rl,
  author    = {Vallat, Gabriel and Wang, Jingwen and Maddux, Anna M. and Kamgarpour, Maryam and Parascho, Stefana},
  title     = {Reinforcement learning for scaffold-free construction of spanning structures},
  booktitle = {Proceedings of the 8th ACM Symposium on Computational Fabrication (SCF)},
  articleno = {12},
  numpages  = {12},
  year      = {2023},
  publisher = {ACM},
  doi       = {10.1145/3623263.3623359},
  address   = {New York, NY, USA}
}

@article{wang2025alternativeplans,
  author  = {Wang, Ziqi and Liu, Wenjun and Wang, Jingwen and Vallat, Gabriel and Shi, Fan and Parascho, Stefana and Kamgarpour, Maryam},
  title   = {Learning to Assemble with Alternative Plans},
  journal = {ACM Transactions on Graphics},
  volume  = {44},
  number  = {4},
  year    = {2025},
  doi     = {10.1145/3730824}
}

@inproceedings{shu2024s2a,
  title     = {Subassembly to Full Assembly: Effective Assembly Sequence Planning through Graph-based Reinforcement Learning},
  author    = {Shu, Chang and Kim, Anton and Park, Shinkyu},
  booktitle = {IEEE International Conference on Robotics and Automation (ICRA)},
  pages     = {14946--14952},
  year      = {2025},
  doi       = {10.1109/ICRA55743.2025.11127710},
  publisher = {IEEE},
  address   = {Atlanta, GA, USA}
}

@misc{tian2025fabrica,
  title         = {Fabrica: Dual-Arm Assembly of General Multi-Part Objects via Integrated Planning and Learning},
  author        = {Tian, Yunsheng and Jacob, Joshua and Huang, Yijiang and Zhao, Jialiang and Gu, Edward and Ma, Pingchuan and Zhang, Annan and Javid, Farhad and Romero, Branden and Chitta, Sachin and Sueda, Shinjiro and Li, Hui and Matusik, Wojciech},
  year          = {2025},
  eprint        = {2506.05168},
  archivePrefix = {arXiv},
  primaryClass  = {cs.RO},
  url           = {https://arxiv.org/abs/2506.05168}
}

@inproceedings{mei2024fabhacks,
  author    = {Mei, Yuxuan and Jones, Benjamin T. and Cascaval, Dan and Mankoff, Jennifer and Vouga, Etienne and Schulz, Adriana},
  title     = {{FabHacks}: Transform Everyday Objects into Home Hacks Leveraging a Solver-aided {DSL}},
  booktitle = {Proceedings of the 9th ACM Symposium on Computational Fabrication (SCF)},
  numpages  = {4},
  year      = {2024},
  publisher = {ACM},
  doi       = {10.1145/3639473.3665788},
  address   = {Aarhus, Denmark}
}

@book{maynard1948mtm,
  author    = {Maynard, Harold B. and Stegemerten, Gustave J. and Schwab, John L.},
  title     = {Methods-Time Measurement},
  publisher = {McGraw-Hill},
  address   = {New York},
  year      = {1948}
}

@article{mazareinezhad2025most,
  author  = {Mazareinezhad, Farbod and Sekkay, Firdaous and Imbeau, Daniel},
  title   = {Improving time estimation accuracy in manufacturing systems: experimental assessment of {MOST} predetermined motion time system using laboratory data and Fitts' law},
  journal = {The International Journal of Advanced Manufacturing Technology},
  volume  = {138},
  pages   = {4721--4737},
  year    = {2025},
  doi     = {10.1007/s00170-025-15791-6}
}

@inproceedings{wilson1996tools,
  author    = {Wilson, Randall H.},
  title     = {A framework for geometric reasoning about tools in assembly},
  booktitle = {Proceedings of the IEEE International Conference on Robotics and Automation (ICRA)},
  pages     = {1837--1844},
  year      = {1996},
  doi       = {10.1109/ROBOT.1996.506979},
  publisher = {IEEE},
  address   = {Minneapolis, MN, USA}
}

@article{kang2008tool,
  author  = {Kang, Xuanyi and Peng, Qingjin},
  title   = {Tool Feasibility Analysis for Fixture Assembly Planning},
  journal = {Journal of Manufacturing Science and Engineering},
  volume  = {130},
  number  = {4},
  pages   = {041010},
  year    = {2008},
  doi     = {10.1115/1.2953072}
}

@article{jones2021automate,
  author  = {Jones, Benjamin and Hildreth, Dalton and Chen, Duowen and Baran, Ilya and Kim, Vladimir G. and Schulz, Adriana},
  title   = {{AutoMate}: A Dataset and Learning Approach for Automatic Mating of {CAD} Assemblies},
  journal = {ACM Transactions on Graphics},
  volume  = {40},
  number  = {6},
  articleno = {227},
  numpages = {18},
  year    = {2021},
  doi     = {10.1145/3478513.3480562}
}

@article{yeo2021feature,
  author  = {Yeo, Changmo and Kim, Byung Chul and Cheon, Sanguk and Lee, Jinwon and Mun, Duhwan},
  title   = {Machining feature recognition based on deep neural networks to support tight integration with {3D} {CAD} systems},
  journal = {Scientific Reports},
  volume  = {11},
  pages   = {22147},
  year    = {2021},
  doi     = {10.1038/s41598-021-01313-3}
}

@inproceedings{sun2023dovetail,
  author    = {Sun, Xingyuan and Cai, Chenyue and Adams, Ryan P. and Rusinkiewicz, Szymon},
  title     = {Gradient-Based Dovetail Joint Shape Optimization for Stiffness},
  booktitle = {Proceedings of the 8th ACM Symposium on Computational Fabrication (SCF)},
  articleno = {5},
  numpages  = {8},
  year      = {2023},
  publisher = {ACM},
  doi       = {10.1145/3623263.3623364},
  address   = {New York, NY, USA}
}

@article{agrawala2003,
  author  = {Agrawala, Maneesh and Phan, Doantam and Heiser, Julie and Haymaker, John and Klingner, Jeff and Hanrahan, Pat and Tversky, Barbara},
  title   = {Designing effective step-by-step assembly instructions},
  journal = {ACM Transactions on Graphics},
  volume  = {22},
  number  = {3},
  pages   = {828--837},
  year    = {2003},
  doi     = {10.1145/882262.882352}
}

@inproceedings{heiser2004,
  author    = {Heiser, Julie and Phan, Doantam and Agrawala, Maneesh and Tversky, Barbara and Hanrahan, Pat},
  title     = {Identification and validation of cognitive design principles for automated generation of assembly instructions},
  booktitle = {Proceedings of the Working Conference on Advanced Visual Interfaces (AVI)},
  pages     = {311--319},
  year      = {2004},
  publisher = {ACM},
  doi       = {10.1145/989863.989917},
  address   = {Gallipoli, Italy}
}

@inproceedings{wang2022ikeamanual,
  author    = {Wang, Ruocheng and Zhang, Yunzhi and Mao, Jiayuan and Zhang, Ran and Cheng, Chin-Yi and Wu, Jiajun},
  title     = {{IKEA-Manual}: Seeing Shape Assembly Step by Step},
  booktitle = {Advances in Neural Information Processing Systems (NeurIPS), Datasets and Benchmarks Track},
  volume    = {35},
  year      = {2022},
  publisher = {Curran Associates, Inc.},
  address   = {New Orleans, LA, USA}
}

@inproceedings{liu2024ikeaatwork,
  author    = {Liu, Yunong and Eyzaguirre, Cristobal and Li, Manling and Khanna, Shubh and Niebles, Juan Carlos and Ravi, Vineeth and Mishra, Saumitra and Liu, Weiyu and Wu, Jiajun},
  title     = {{IKEA} Manuals at Work: {4D} Grounding of Assembly Instructions on Internet Videos},
  booktitle = {Advances in Neural Information Processing Systems (NeurIPS), Datasets and Benchmarks Track},
  volume    = {37},
  year      = {2024},
  publisher = {Curran Associates, Inc.},
  address   = {Vancouver, BC, Canada}
}

@inproceedings{wang2022lego,
  author    = {Wang, Ruocheng and Zhang, Yunzhi and Mao, Jiayuan and Cheng, Chin-Yi and Wu, Jiajun},
  title     = {Translating a Visual {LEGO} Manual to a Machine-Executable Plan},
  booktitle = {Computer Vision -- ECCV 2022},
  pages     = {677--694},
  year      = {2022},
  publisher = {Springer},
  doi       = {10.1007/978-3-031-19836-6\_38},
  address   = {Cham, Switzerland}
}

@inproceedings{tie2025manual2skill,
  author    = {Tie, Chenrui and Sun, Shengxiang and Zhu, Jinxuan and Liu, Yiwei and Guo, Jingxiang and Hu, Yue and Chen, Haonan and Chen, Junting and Wu, Ruihai and Shao, Lin},
  title     = {{Manual2Skill}: Learning to Read Manuals and Acquire Robotic Skills for Furniture Assembly Using Vision-Language Models},
  booktitle = {Robotics: Science and Systems (RSS)},
  year      = {2025},
  publisher = {Robotics: Science and Systems Foundation},
  address   = {Los Angeles, CA, USA}
}

@inproceedings{zhang2025manualpa,
  author    = {Zhang, Jiahao and Cherian, Anoop and Rodriguez, Cristian and Deng, Weijian and Gould, Stephen},
  title     = {{Manual-PA}: Learning {3D} Part Assembly from Instruction Diagrams},
  booktitle = {Proceedings of the IEEE/CVF International Conference on Computer Vision (ICCV)},
  year      = {2025},
  publisher = {IEEE},
  address   = {Honolulu, HI, USA}
}

@inproceedings{parzeller2026aissembly,
  author    = {Parzeller, Rafael and Robl, Peter and Gerhard, Detlef},
  title     = {{AIssembly}: Generating Assembly Work Instructions from {3D-CAD} Models Using Artificial Intelligence},
  booktitle = {Product Lifecycle Management. {PLM} in the Age of Model-Based Engineering in Industry (PLM 2025)},
  series    = {IFIP Advances in Information and Communication Technology},
  volume    = {772},
  year      = {2026},
  publisher = {Springer},
  doi       = {10.1007/978-3-032-09700-2\_20},
  address   = {Cham, Switzerland}
}

@inproceedings{widulle2024instructions,
  author    = {Widulle, Niklas and Meyer, Fr{\'e}d{\'e}ric and Niggemann, Oliver},
  title     = {Generating Assembly Instructions Using Reinforcement Learning in Combination with Large Language Models},
  booktitle = {IEEE 22nd International Conference on Industrial Informatics (INDIN)},
  pages     = {1--7},
  year      = {2024},
  doi       = {10.1109/INDIN58382.2024.10774545},
  publisher = {IEEE},
  address   = {Beijing, China}
}

@article{marino2024ar,
  author  = {Marino, Emanuele and Barbieri, Loris and Bruno, Fabio and Muzzupappa, Maurizio},
  title   = {Assessing user performance in augmented reality assembly guidance for industry 4.0 operators},
  journal = {Computers in Industry},
  volume  = {157},
  pages   = {104085},
  year    = {2024},
  doi     = {10.1016/j.compind.2024.104085}
}

@inproceedings{moisi2025digitaltwins,
  author    = {Moisi, Alexandra and Kenny, Joseph Clair and Bucklin, Oliver and Aejmelaeus-Lindstr{\"o}m, Petrus and Kango, Ananya and Casas, Gonzalo and Mitterberger, Daniela and Gramazio, Fabio and Kohler, Matthias},
  title     = {Interactive Digital Twins: Integrating {XR} and Motion Capture for Timber Assembly},
  booktitle = {Proceedings of the ACM Symposium on Computational Fabrication (SCF)},
  articleno = {3},
  numpages  = {18},
  year      = {2025},
  publisher = {ACM},
  doi       = {10.1145/3745778.3766661}
}

@misc{jia2026omnispatial,
  title         = {{OmniSpatial}: Towards Comprehensive Spatial Reasoning Benchmark for Vision Language Models},
  author        = {Jia, Mengdi and Qi, Zekun and Zhang, Shaochen and Zhang, Wenyao and Yu, Xinqiang and He, Jiawei and Wang, He and Yi, Li},
  year          = {2026},
  eprint        = {2506.03135},
  archivePrefix = {arXiv},
  primaryClass  = {cs.CV},
  url           = {https://arxiv.org/abs/2506.03135}
}

@inproceedings{khan2024text2cad,
  author    = {Khan, Mohammad Sadil and Sinha, Sankalp and Sheikh, Talha Uddin and Stricker, Didier and Ali, Sk Aziz and Afzal, Muhammad Zeshan},
  title     = {{Text2CAD}: Generating Sequential {CAD} Models from Beginner-to-Expert Level Text Prompts},
  booktitle = {Advances in Neural Information Processing Systems (NeurIPS)},
  volume    = {37},
  year      = {2024},
  publisher = {Curran Associates, Inc.},
  address   = {Vancouver, BC, Canada}
}

@misc{berger2026physicsinloop,
  title         = {Physics-in-the-Loop: A Hybrid Agentic Architecture for Validated {CAD} Engineering Design},
  author        = {Berger, Elias and Usama, Muhammad and Mehlst{\"a}ubl, Jan and Saske, Bernhard and Paetzold-Byhain, Kristin},
  year          = {2026},
  eprint        = {2605.19717},
  archivePrefix = {arXiv},
  primaryClass  = {cs.AI},
  url           = {https://arxiv.org/abs/2605.19717}
}

@article{wan2018assembly,
  title={Assembly sequence planning for motion planning},
  author={Wan, Weiwei and Harada, Kensuke and Nagata, Kazuyuki},
  journal={Assembly Automation},
  volume={38},
  number={2},
  pages={195--206},
  year={2018},
  publisher={Emerald Publishing Limited}
}

@inproceedings{spyridi1990accessibility,
    title = {Accessibility analysis for the automatic inspection of mechanical parts by coordinate measuring machines},
    author = {Antonios J. Spyridi and Aristides A. G. Requicha},
    booktitle = {Proceedings of the IEEE International Conference on Robotics and Automation (ICRA)},
    pages = {1284--1289},
    year = {1990},
    doi = {10.1109/ROBOT.1990.126176},
    publisher = {IEEE},
    address   = {Cincinnati, OH, USA}
}

@article{spitz1999accessibility,
    title = {Accessibility analysis for planning of dimensional inspection with coordinate measuring machines},
    author = {Steven N. Spitz and Antonios J. Spyridi and Aristides A. G. Requicha},
    journal = {IEEE Transactions on Robotics and Automation},
    volume = {15},
    number = {4},
    pages = {714--727},
    year = {1999}
}

@article{spitz2000graphics,
    title = {Accessibility analysis using computer graphics hardware},
    author = {Steven N. Spitz and Aristides A. G. Requicha},
    journal = {IEEE Transactions on Visualization and Computer Graphics},
    volume = {6},
    number = {3},
    pages = {208--219},
    year = {2000}
}

@article{dhaliwal2003cones,
    title = {Algorithms for computing global accessibility cones},
    author = {Savinder Dhaliwal and Satyandra K. Gupta and Jun Huang and Alok Priyadarshi},
    journal = {Journal of Computing and Information Science in Engineering},
    volume = {3},
    number = {3},
    pages = {200--209},
    year = {2003},
    doi = {10.1115/1.1606475}
}

@article{chen1993parting,
    title = {Parting directions for mould and die design},
    author = {Lin-Lin Chen and Shuo-Yan Chou and Tony C. Woo},
    journal = {Computer-Aided Design},
    volume = {25},
    number = {12},
    pages = {762--768},
    year = {1993},
    doi = {10.1016/0010-4485(93)90103-U}
}

@article{khardekar2006mold,
    title = {Finding feasible mold parting directions using graphics hardware},
    author = {Rahul Khardekar and Greg Burton and Sara McMains},
    journal = {Computer-Aided Design},
    volume = {38},
    number = {4},
    pages = {327--341},
    year = {2006}
}

@article{elber1994milling,
    title = {Accessibility in 5-axis milling environment},
    author = {Gershon Elber},
    journal = {Computer-Aided Design},
    volume = {26},
    number = {11},
    pages = {796--802},
    year = {1994}
}

@article{ahn2002cast,
    title = {Separating an object from its cast},
    author = {Hee-Kap Ahn and Mark de Berg and Prosenjit Bose and Siu-Wing Cheng and Dan Halperin and Ji{\v{r}}{\'\i} Matou{\v{s}}ek and Otfried Schwarzkopf},
    journal = {Computer-Aided Design},
    volume = {34},
    number = {8},
    pages = {547--559},
    year = {2002}
}

@inproceedings{zhong2025deepmill,
    title = {{DeepMill}: Neural Accessibility Learning for Subtractive Manufacturing},
    author = {Fanchao Zhong and Yang Wang and Peng-Shuai Wang and Lin Lu and Haisen Zhao},
    booktitle = {Proceedings of the Special Interest Group on Computer Graphics and Interactive Techniques Conference Conference Papers (SIGGRAPH Conference Papers '25)},
    year = {2025},
    publisher = {ACM},
    doi = {10.1145/3721238.3730657},
    address   = {Vancouver, BC, Canada}
}

@article{wang2023frameassembly,
    title = {A Temporal Coherent Topology Optimization Approach for Assembly Planning of Bespoke Frame Structures},
    author = {Ziqi Wang and Florian Kennel-Maushart and Yijiang Huang and Bernhard Thomaszewski and Stelian Coros},
    journal = {ACM Transactions on Graphics},
    volume = {42},
    number = {4},
    pages = {1--13},
    year = {2023},
    doi = {10.1145/3592102}
}

@article{huang2024constructability,
    title = {Constructability-driven design of frame structures with state-space search methods},
    author = {Yijiang Huang and Caelan Reed Garrett and Caitlin T. Mueller},
    journal = {Automation in Construction},
    volume = {167},
    pages = {105711},
    year = {2024},
    doi = {10.1016/j.autcon.2024.105711}
}

\appendix
\section{Supplementary Material}

\subsection{Contact-Point and Axis Selection}
\label{sec:appendix:contact}
\begin{figure*}[htbp]
  \centering
  \begin{minipage}{0.48\textwidth}
    \centering
    \includegraphics[width=\linewidth]{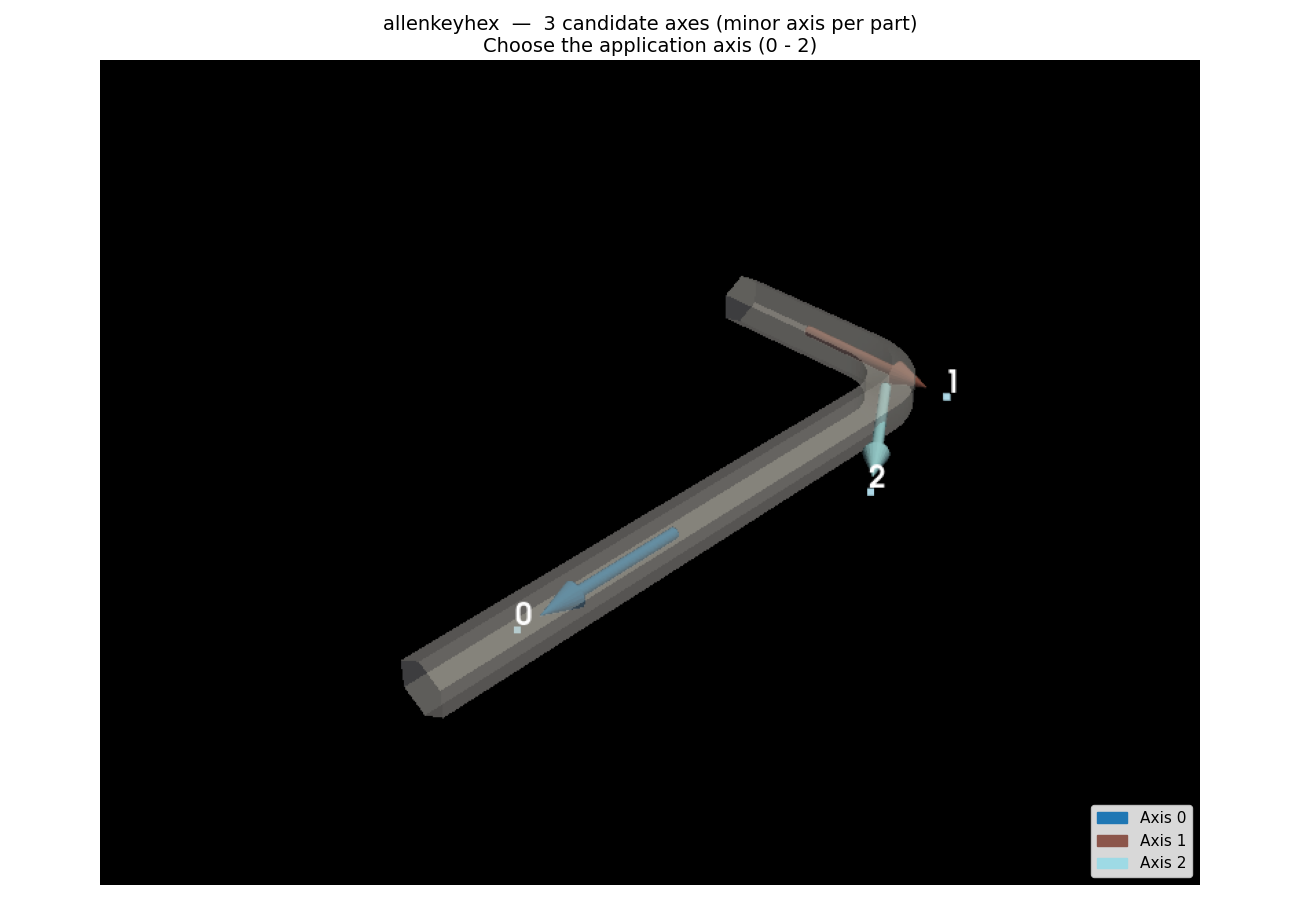}\\
    {\small (a) Axis selection}
  \end{minipage}\hfill
  \begin{minipage}{0.48\textwidth}
    \centering
    \includegraphics[width=\linewidth]{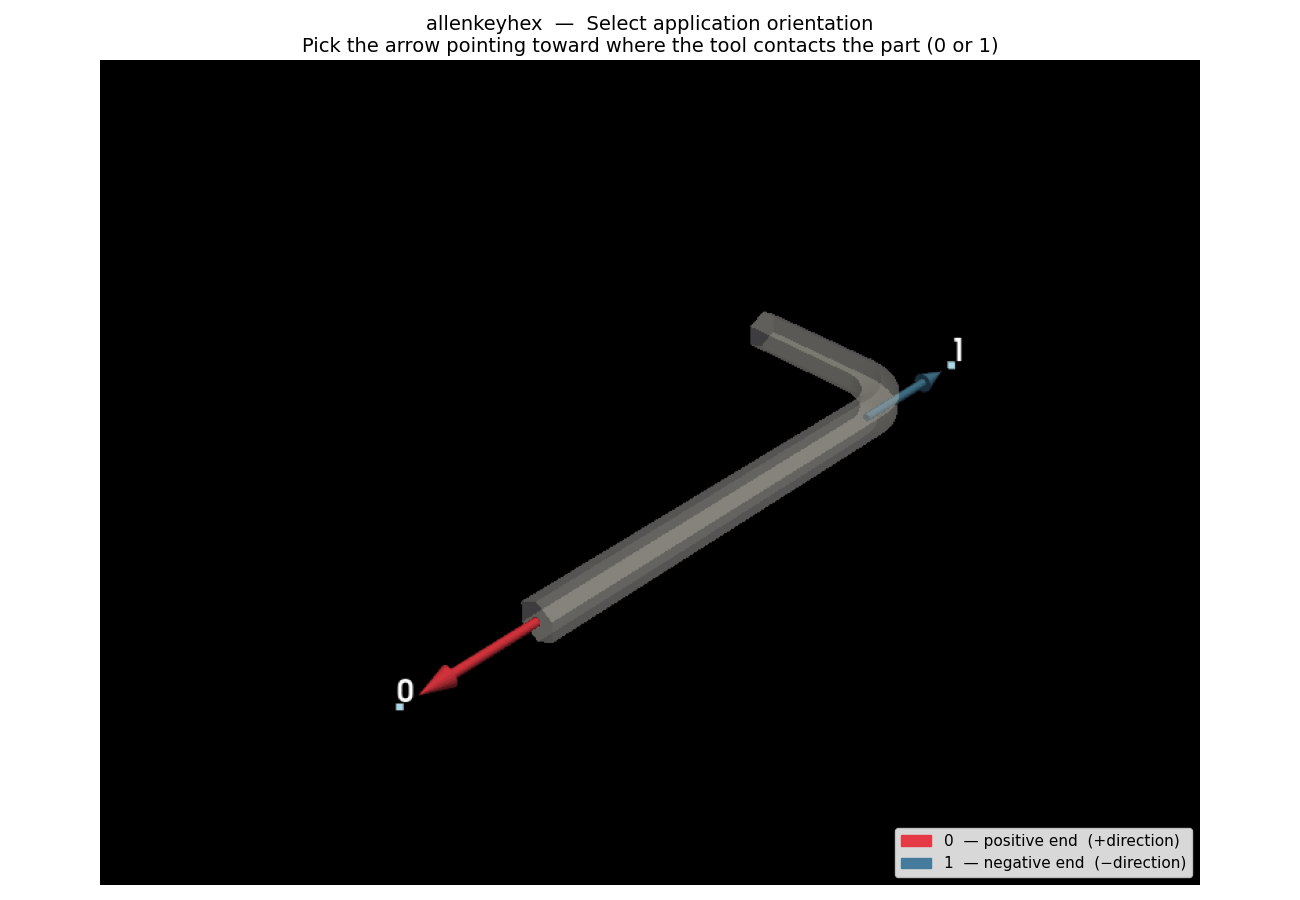}\\
    {\small (b) Direction and contact point}
  \end{minipage}
  \caption{The two renders shown to the VLM when recovering the application axis
  of a tool, here an Allen key (Section~\ref{sec:methods:tools}). (a) The tool
  mesh is split into three convex components, and the minor principal axis of
  each is overlaid as a numbered candidate; the VLM picks the application axis.
  (b) Given the chosen axis, the two opposite directions along it are drawn, and
  the VLM picks the one pointing towards where the tool contacts the part, which
  fixes the contact point and application direction.}
  \Description{Two renders of a translucent grey L-shaped Allen key on a black
  background. In the left render, three coloured arrows labelled 0, 1, and 2 lie
  along the long arm, the short arm, and the bend, each marking the minor
  principal axis of one convex component. In the right render, two arrows point
  in opposite directions along the long arm, a red one labelled 0 at the free end
  of the long arm and a blue one labelled 1 at the bend.}
  \label{fig:appendix:tool_axis}
\end{figure*}

Figure~\ref{fig:appendix:tool_axis} shows the two renders presented to the VLM
during axis and direction selection (Section~\ref{sec:methods:tools}): the
candidate axes from the convex decomposition in a first pass, and the two
directions along the chosen axis in a second.

For the contact-point and contact-axis selection, we use 10 screwing tools
from GrabCAD as a dataset. For each tool we label the
correct application axis and direction by hand and run the selection 10 times to
check the VLM's consistency, clustering the ten picks by direction so that each
distinct answer is validated against the hand-labelled ground truth only once.
Every tool's ten runs collapse to a single cluster, confirming that the VLM is
internally consistent, and it selects both the correct axis and direction on 5
of the 10 tools. The errors split roughly equally
between cases where the VLM is given a correct candidate set and still chooses
wrongly, and cases where the convex-decomposition step that produces the
candidates (Section~\ref{sec:methods:tools}) fails to surface the correct axis,
leaving the VLM with no correct option to pick. The two error types fall on
different kinds of tool: the decomposition failures on wrenches, and the VLM
errors on tools presenting many candidate axes, such as electric drills.

\subsection{Directional Blocking Matrix}
\label{sec:appendix:blocking_matrix}
\begin{figure*}[htbp]
  \centering
  \includegraphics[width=\linewidth]{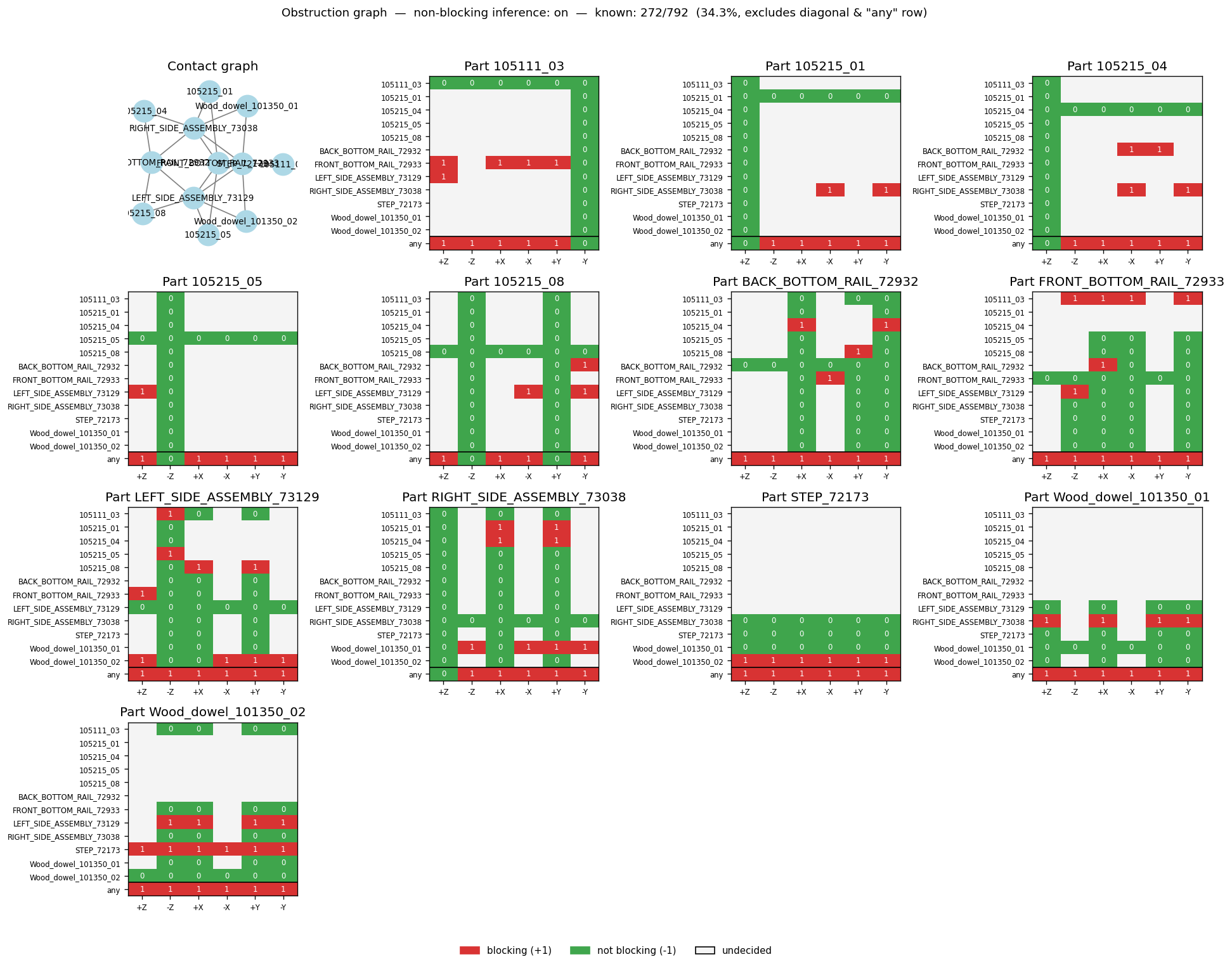}
  \caption{Approximate directional blocking matrix $B$ and contact graph for one
  IKEA assembly (Section~\ref{sec:methods:subassembly}). Top-left: the contact
  graph, whose nodes are parts and whose edges join parts in surface contact. Each
  remaining panel is the blocking matrix of a single part: its rows are the other
  parts, its columns are the six axis-aligned directions ($\pm x$, $\pm y$,
  $\pm z$), and each cell is coloured red where that part blocks the panel's part
  in that direction, green where it does not, and white where the relation is
  still undecided. The bottom \texttt{any} row summarises, per direction, whether
  the part is blocked by any other part. Only the entries observed during the
  disassembly-path search are decided (here 272 of 792), yet this partial matrix
  already suffices to enumerate candidate subassemblies.}
  \Description{A grid of small heatmaps, one per part of an IKEA assembly, with a
  node-link contact graph in the top-left corner. In each heatmap, rows are the
  other parts and columns are the six axis-aligned directions; cells are red for
  blocking, green for non-blocking, and white for undecided, with a summary row at
  the bottom.}
  \label{fig:appendix:blocking_matrix}
\end{figure*}

Figure~\ref{fig:appendix:blocking_matrix} shows the approximate directional
blocking matrix $B$ of Section~\ref{sec:methods:subassembly}, together with the
contact graph it is derived from, for one IKEA assembly. Because $B$ is populated
incrementally as a by-product of the disassembly-path search, only a fraction of
its entries are decided (here 272 of 792), which nonetheless suffices to
enumerate candidate subassemblies without additional simulation.

\subsection{Beam Search Approach}
\label{sec:appendix:beam_search}
\begin{figure}[H]
  \centering
  \includegraphics[width=\linewidth]{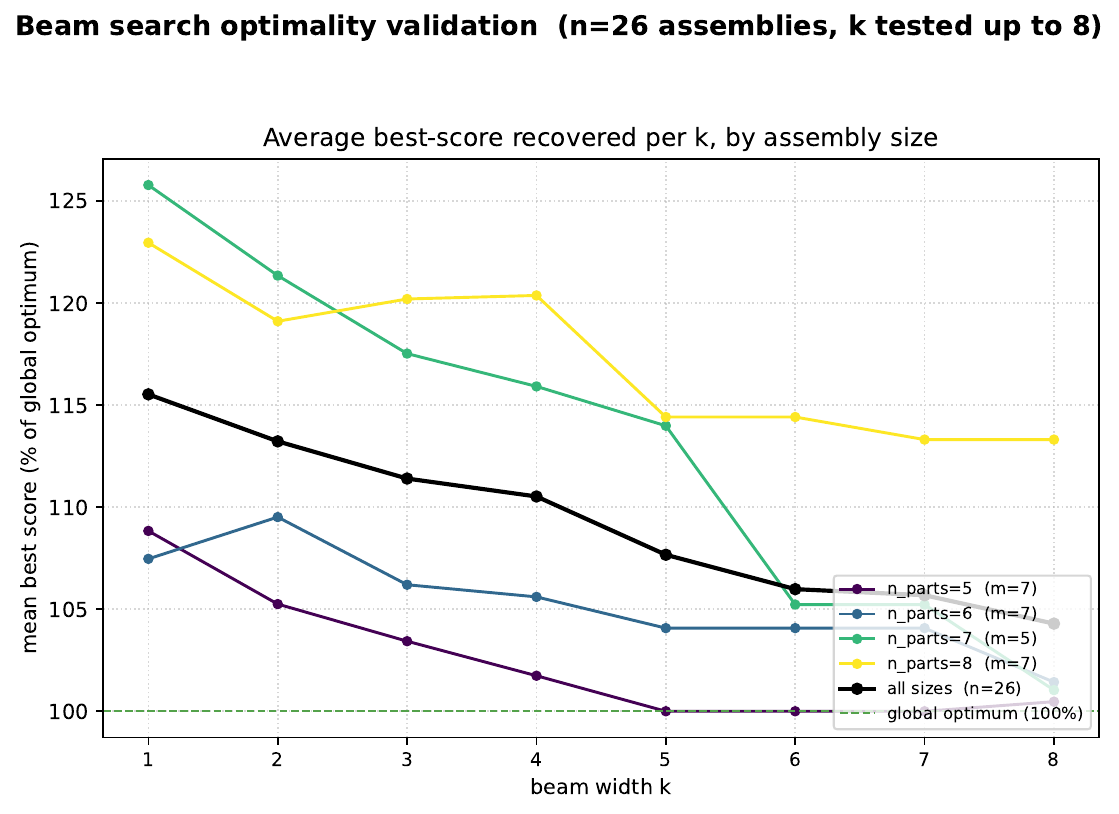}
  \caption{Cost of the best sequence at different widths $k$, for assemblies with
  up to 8 parts. The plots show the average cost relative to the global optimum
  as a function of $k$. In 15 of the 26 assemblies, the global optimum is
  recovered at $k \leq n_i$, where $n_i$ is the number of parts in assembly $i$.}
  \Description{Line plot of average sequence cost relative to the global optimum
  against beam width k for assemblies of up to eight parts. The cost falls
  roughly linearly from about 115 percent at k equals one to about 104 percent
  at k equals eight.}
  \label{fig:appendix:beam_search}
\end{figure}

To check how close the beam search comes to the global optimum of the cost
function, an exhaustive search over the feasibility tree is run separately to
enumerate every feasible sequence and record the lowest-cost one. The beam
search is then run at each integer beam width $k$, and the best sequence it finds
is compared against the exhaustive optimum. Figure~\ref{fig:appendix:beam_search}
shows the cumulative cost $J$ of the beam-search result as a function of $k$. The
study uses 26 assemblies sampled at random from the test split of the ASAP
dataset (Section~\ref{sec:methods:experimental_evaluation}), so they may also
appear in the other experiments. The number of parts is
limited to 8, since an exhaustive search of larger
assemblies becomes exponentially more time consuming, and the sample size is
large enough to show a general trend in the data.

The global average cost decreases roughly linearly with the beam width, moving
from 115.5\% at $k=1$ to 104.3\% at $k=8$. As expected, the
overall cost overhead increases with the number of parts in the assembly.

Narrow beam widths are therefore sufficient in the common case, and the tunable
$k$ trades compute for sequence quality: small $k$ for a tight budget, larger $k$
when a lower-cost sequence is worth the extra simulation, with $k$ needing to
scale with the number of parts to hold the same relative performance. A nearly
optimal sequence can thus be found with a small horizon, while the true optimum
requires an exhaustive search; humans appear to sequence assemblies with a
similarly bounded horizon, weighing immediate against anticipated future planning
cost rather than optimising globally~\cite{binder2025subgoals}.

\subsection{Sequence-Planning Runtime}
\label{sec:appendix:runtime}
\begin{figure}[H]
  \centering
  \includegraphics[width=0.95\linewidth]{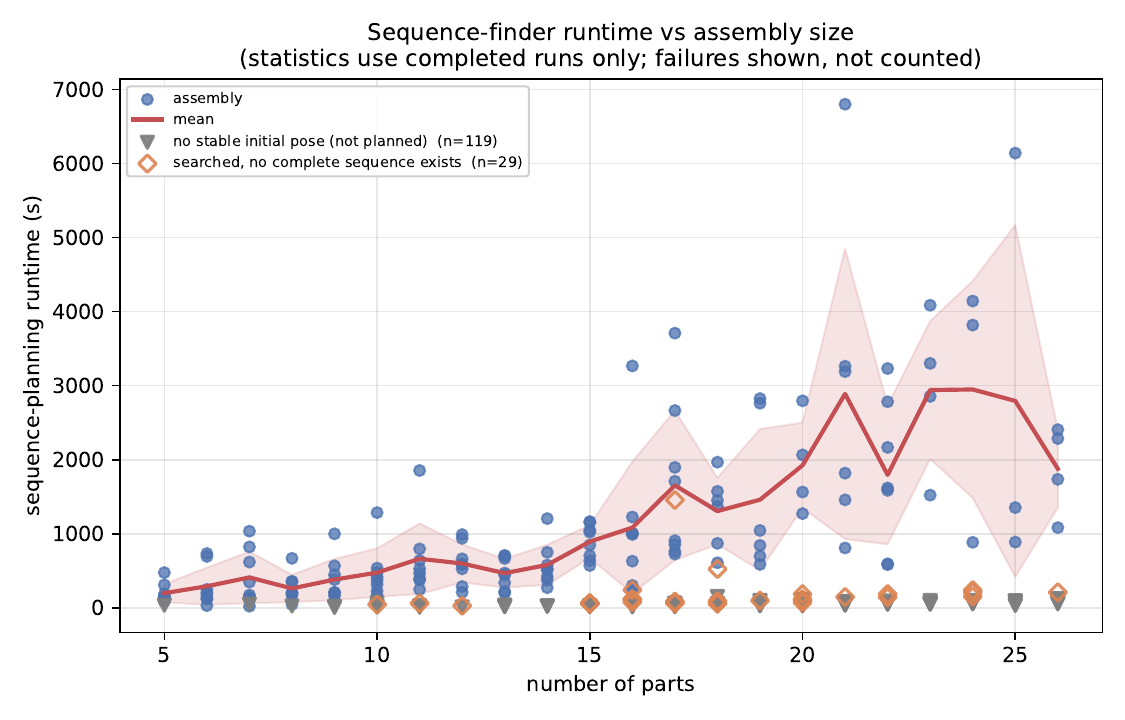}
  \caption{Sequence-planning runtime against the number of parts, for 295
  assemblies from the ASAP dataset ($k = 4$), without
  subassembly detection or tool checking. Blue dots are completed runs, the red
  line is their mean per part count, and the shaded band shows one standard
  deviation. Assemblies without a
  stable initial pose (grey triangles, $n=119$) and searches that find no
  complete sequence (orange diamonds, $n=29$) are shown but not included in the
  statistics. Up to 17 parts, each part count has 8 completed runs; above 17,
  4.78 on average.}
  \Description{Scatter plot of sequence-planning runtime in seconds, from 0 to
  7000, against the number of parts, from 5 to 26. Blue dots mark completed runs
  and a red line their mean, with a light red band of one standard deviation
  around it. The mean stays
  below about 700 seconds up to 14 parts, rises to about 1600 seconds at 17
  parts and varies between about 1800 and 3000 seconds from 21 to 25 parts, with
  a wide band at the larger sizes; single runs reach about 6800 and 6100 seconds
  at 21 and 25 parts.
  Grey triangles and orange diamonds near zero mark failed runs, which are not
  counted in the mean.}
  \label{fig:appendix:runtime}
\end{figure}

Figure~\ref{fig:appendix:runtime} shows the runtime of the sequence planner
(Section~\ref{sec:methods:sequencing}) as a function of the number of parts, on
295 assemblies of 5 to 26 parts from the ASAP dataset
(Section~\ref{sec:methods:experimental_evaluation}). The assemblies are taken
from the test split first; for part counts where it holds too few, further
assemblies are drawn at random from the training split. Each run covers the full linear sequence search with
beam width $k = 4$ (Section~\ref{sec:methods:search}), without
subassembly detection (Section~\ref{sec:methods:subassembly}) and without tool
checking, which is disabled for the reason given in
Section~\ref{sec:methods:experimental_evaluation}. All runs are made on a single
machine with an AMD EPYC 9454P processor (48 cores, 96 logical processors,
2.75\,GHz) and 128\,GB of RAM at 4800\,MT/s.

Of the 295 assemblies, 148 do not yield a sequence: 119 have no stable initial
pose and are not planned, and 29 are searched without a complete sequence being
found. These runs are shown in the figure but excluded from the runtime
statistics. The remaining 147 runs give 8 completed sequences for every part
count up to 17, and 4.78 per part count on average above 17. The sample is smaller for
larger assemblies for two reasons: the dataset contains fewer of them, and all
available assemblies above 17 parts are used; and a larger share of the runs
fail, since every additional part adds a chance that the assembly or one of its
partial states is unstable.

The mean time per part grows from 39.7\,s at 5 parts to 107.4\,s at 21 to 26
parts. In total, a plan takes 199\,s on average at 5 parts (median 146\,s) and
2{,}560\,s over the 23 completed runs with 21 to 25 parts (median 2{,}166\,s). Most of this
increase comes from the stability check. It simulates all parts remaining in the
partial assembly together under gravity, in each of three resting poses
(Section~\ref{sec:methods:feasibility}), so its cost grows with the number of
parts. Path finding stays roughly constant across assembly sizes: the parts that
are not being removed can be treated as a single fixed body, so only two bodies
are simulated at a time, whatever the size of the assembly. The spread between
assemblies of the same size also widens with the number of parts; the slowest
completed run, an assembly of 21 parts, takes 6{,}799\,s.

\subsection{Sequence Planner Pseudocode}
\label{sec:appendix:sequence_planner}
Algorithm~\ref{alg:appendix:sequence_planner} summarises the sequence planner of Section~\ref{sec:methods:sequencing}. Phase~1 is the parallelised beam search introduced in Sections~\ref{sec:methods:feasibility} and~\ref{sec:methods:search}, which also accumulates the directional blocking matrix $B$ as a by-product of the per-step path planning. Phase~2 is the subassembly detection of Section~\ref{sec:methods:subassembly}, which operates on the linear sequence $\sigma$ produced by Phase~1 and only consults $B$ together with the physics simulator.

\begin{algorithm}[htbp]
\SetAlgoLined
\KwIn{mesh assembly $A$, tool catalogue $T$, beam width $k$, subassembly width $k_S$}
\KwOut{assembly sequence $\sigma$, hierarchical decomposition $\mathcal{S}$}
\BlankLine
\tcc{Phase 1: parallelised beam search over part removals}
$F \gets \{(\emptyset,\, A,\, 0)\}$ \tcp*{frontier: (partial sequence, remaining parts, cost)}
$B \gets$ empty directional blocking matrix\;
\While{any entry of $F$ has remaining parts}{
  $C \gets \emptyset$\;
  \ForEach(\tcp*[h]{in parallel}){$(s, A_s, J_s) \in F$ with $A_s \neq \emptyset$}{
    \ForEach{$p \in A_s$}{
      \If{$p$ is feasibly removable from $A_s$ \tcp*[r]{path, tool, stability}}{
        Update $B$ with per-direction blocking observed during the check\;
        Append $(s \cdot p,\, A_s \setminus \{p\},\, J_s + C(p))$ to $C$\;
      }
    }
  }
  $F \gets$ the $k$ entries of $C$ with the lowest cumulative cost\;
}
$\sigma \gets$ lowest-cost complete sequence over all removal steps checked so far (shortest path, no further simulation)\;
\BlankLine
\tcc{Phase 2: subassembly detection on $\sigma$}
$\mathcal{S} \gets \{\sigma\}$\;
\ForEach{segment $\tau \in \mathcal{S}$ that has not yet been split}{
  Enumerate connected $S \subseteq \tau$ by tree search over $B$, recording $S$ as a candidate when at least one direction $\mathbf{d}$ has external blocking $c_\mathbf{d}(S) = 0$\;
  Score each candidate by size, freeness, and remainder fragmentation\;
  \ForEach{top-$k_S$ candidate $S$ by score}{
    \If{$S$ is removable in the physics simulator and $\mathrm{cost}(\{S, \tau\setminus S\}) < \mathrm{cost}(\tau)$}{
      Replace $\tau$ in $\mathcal{S}$ by $\{S,\, \tau \setminus S\}$ and recurse on both\;
      \textbf{break}\;
    }
  }
}
\Return $(\sigma, \mathcal{S})$\;
\caption{Sequence planner with subassembly detection.}
\label{alg:appendix:sequence_planner}
\end{algorithm}

\subsection{Design Feedback Prompts}
\label{sec:appendix:feedback_prompt}
The two kinds of feedback in Section~\ref{sec:methods:feedback} use separate
prompts, each passed to the VLM together with the structured record of the step
or problem and its renders.

For assemblies that can be built (Section~\ref{sec:methods:feedback}), the VLM
reviews one step at a time with the following prompt, which lists five common
assembly pitfalls and returns feedback only when a step warrants it:

\begin{quote}\small\raggedright
You are an expert design engineer specialising in Design for Assembly (DFA) and
human ergonomics. You review ONE assembly step and produce HIGH-SIGNAL, SELECTIVE
feedback.

Rules:
\begin{enumerate}
  \item Each feedback point must be NON-TRIVIAL and specific to this part's
    geometry or this step's situation. Skip generic DFA advice and self-evident
    remarks.
  \item If nothing actionable stands out, set \texttt{needs\_feedback=\allowbreak false} and
    leave the points list empty. Do NOT pad with weak content.
  \item Categorise each point as `observation' (something specific in the
    images), `dfa\_issue' (a real DFA pitfall present here), or `recommendation'
    (a concrete geometric change). Skip categories that don't apply.
  \item Reference visible part features or state numbers. Don't restate
    definitions.
\end{enumerate}

\textbf{DFA pitfalls to consider (cite only when actually present)}
\begin{enumerate}
  \item Lack of clearance and access: blind insertions, no visual access to
    mating surfaces.
  \item Poor alignment features: flat-on-flat insertions, missing
    chamfers/tapers/pins.
  \item Ergonomic strain: awkward wrist angles, multi-axis simultaneous
    alignment.
  \item Ambiguous orientation: slight asymmetries that allow incorrect
    installation.
  \item Lack of self-retention: parts that fall before fastening.
\end{enumerate}

\end{quote}

For assemblies that cannot be built, including the examples in
Section~\ref{sec:results:feedback}, an earlier version of the prompt gave the
model a longer list of assembly guidelines and asked for categorised feedback.
We condensed it to the version below, which gave better results in our runs and
keeps the wording plain enough for a designer who is not an assembly specialist:

\begin{quote}\small
You explain to the designer of a product why one step of putting it together
does not work, and suggest how to change a part so that it does. The designer
knows the product well but is not an assembly specialist. Write the way a
helpful colleague would say it out loud: plain words, short sentences.

Write two or three sentences:
\begin{enumerate}
  \item What goes wrong, as a person holding the parts would notice it (e.g.\
    `Until the lid is on, nothing stops the two hinge pins sliding out of their
    holes.').
  \item The simplest change to one part's shape that fixes it, and which part to
    change. Suggest one fix, not a combination. Add a second option only if it
    is clearly worth knowing, in one short sentence.
\end{enumerate}

Rules:
\begin{itemize}
  \item Name parts only by the names you are given. When two parts share a name,
    say `both pivot link plates' or `the three spacers', not the name twice.
  \item Never mention colours, part numbers, image views, the planner or the
    simulation. The text may be read without the picture.
  \item Use everyday words. Common CAD words such as hole, slot, pin, tab, lip or
    chamfer are fine. Avoid specialist terms such as self-retaining, captive,
    axially, annular, circumferential, detent, interlock or mating; say what the
    feature does instead (`a small lip that stops the plate sliding off the
    pin').
  \item Keep each part's job in mind: a part that must pivot or slide still has
    to after your change.
  \item Only claim what the picture and the facts you are given show. If they do
    not show why it fails, say plainly what is known and keep the fix general.
  \item Suggest changes to part shapes only, never to the order of the steps: no
    order works with the current shapes.
  \item No headings, lists or preamble.
\end{itemize}
\end{quote}

\subsection{Example Manual Page}
\label{sec:appendix:manual_example}
\begin{figure*}[p]
    \centering
    \includegraphics[width=0.9\linewidth,height=0.85\textheight,keepaspectratio]{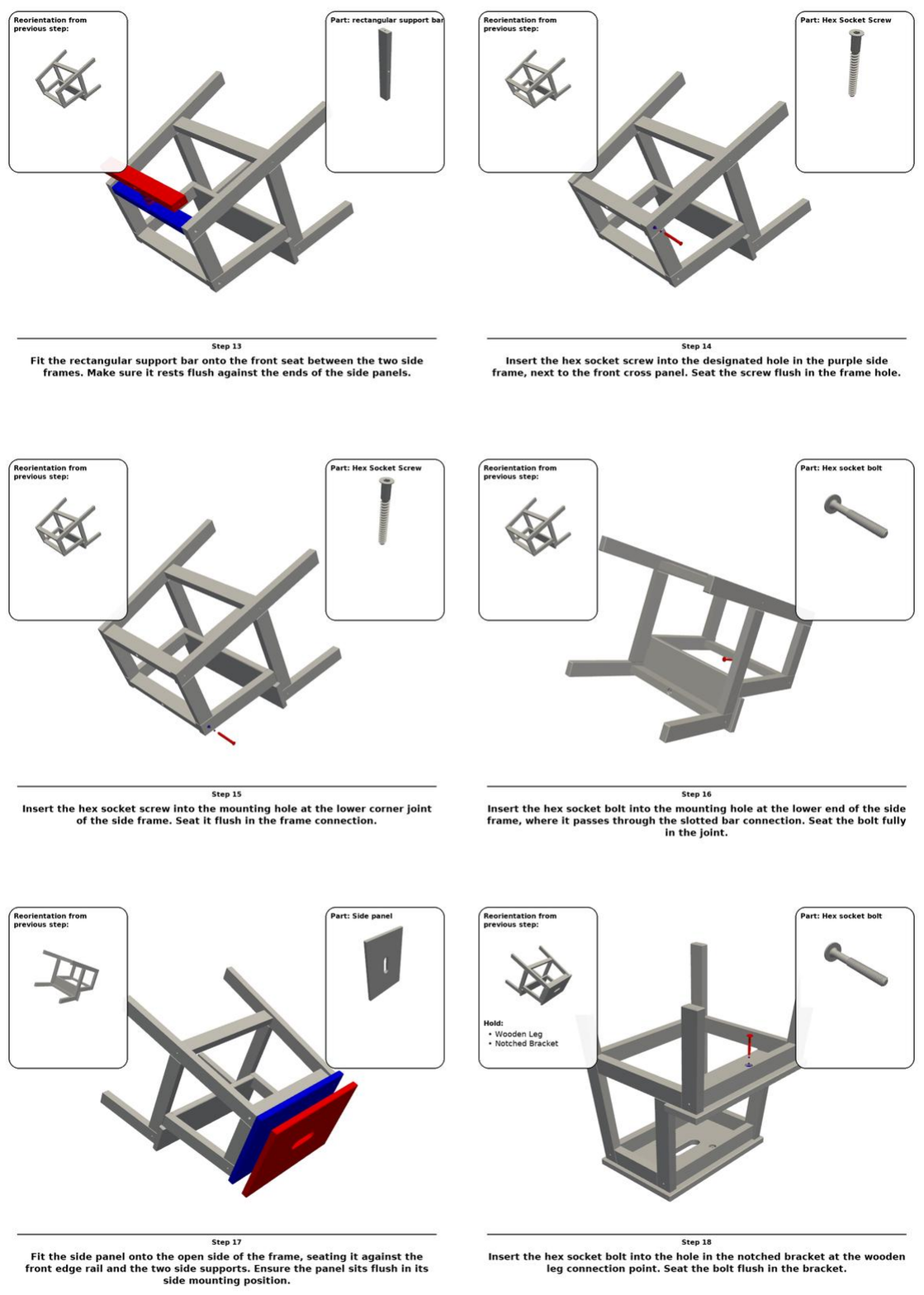}
  \caption{An example page from a generated assembly manual consisting of six steps. Each panel combines a
  coloured render of the action, with the part being added highlighted against the
  rest of the assembly and its motion marked, a text instruction, and conditional
  callouts identifying the part being added and any required reorientation
  relative to the previous step.}
  \Description{One manual page holding six numbered assembly steps, laid out as
  three rows of two panels. Each panel shows a large render of the step, with the
  part being added highlighted in colour against the grey assembly and its
  insertion direction marked by an arrow, above a numbered caption giving the
  instruction in words. A small boxed render in the upper left of each panel
  shows the reorientation from the previous step, and a second box in the upper
  right names and depicts the part being added.}
  \label{fig:appendix:manual_example}
\end{figure*}

Figure~\ref{fig:appendix:manual_example} shows an example page produced by the
manual generator of Section~\ref{sec:methods:manual}, covering six assembly steps.

\end{document}